\documentclass[nohyperref]{ieeeaccess}
\usepackage{cite}
\usepackage[dvipsnames]{xcolor}
\usepackage{amsmath,amssymb,amsfonts}
\usepackage{algorithm}
\usepackage{algorithmic}
\usepackage{graphicx}
\usepackage{textcomp}
\usepackage{booktabs}
\usepackage{array}

\usepackage{booktabs}      
\usepackage{enumitem}
\usepackage{multirow}
\usepackage{graphicx}
\usepackage{varioref}
\usepackage{subcaption}

\usepackage{hyperref}
\usepackage{color}

\definecolor{Red}{rgb}{0.6,0,0}
\definecolor{Blue}{rgb}{0,0,0.8}
\definecolor{Green}{rgb}{0,0.7,0.3}
\definecolor{airforceblue}{rgb}{0.36, 0.54, 0.66}
\definecolor{ao(english)}{rgb}{0.0, 0.5, 0.0}
\definecolor{azure(colorwheel)}{rgb}{0.0, 0.5, 1.0}
\definecolor{crimson}{rgb}{0.86, 0.08, 0.24}
\definecolor{darkcerulean}{rgb}{0.03, 0.27, 0.49}
\definecolor{cobalt}{rgb}{0.0, 0.28, 0.67}
\definecolor{rosegold}{rgb}{0.72, 0.43, 0.47}
\definecolor{orange-red}{rgb}{1.0, 0.27, 0.0}
\definecolor{mountainmeadow}{rgb}{0.19, 0.73, 0.56}
\definecolor{malachite}{rgb}{0.04, 0.85, 0.32}
\definecolor{darkblue}{rgb}{0.0, 0.0, 0.55}
\definecolor{customblue}{rgb}{0.2, 0.35, 0.8}
\definecolor{gg}{gray}{0.92}
\newcolumntype{a}{>{\columncolor{gg}}c}

\definecolor{gg}{gray}{0.9}

\hypersetup{colorlinks=true}
\hypersetup{linktoc=all}
\hypersetup{citecolor=customblue}
\hypersetup{linkcolor=crimson}
\hypersetup{urlcolor=darkblue}

\newcommand{\ie}{\textit{i.e.}}
\newcommand{\eg}{\textit{e.g.}}
\newcommand{\mujoco}{MuJoCo }

\newcommand{\specialcell}[2][c]{
  \begin{tabular}[#1]{@{}c@{}}#2\end{tabular}}

\newtheorem{prop}{\textbf{Proposition}}

\newcommand{\PreserveBackslash}[1]{\let\temp=\\#1\let\\=\temp}
\newcolumntype{C}[1]{>{\centering\arraybackslash}p{#1}}

\def\BibTeX{{\rm B\kern-.05em{\sc i\kern-.025em b}\kern-.08em
    T\kern-.1667em\lower.7ex\hbox{E}\kern-.125emX}}

\begin{document}
\history{Date of publication xxxx 00, 0000, date of current version xxxx 00, 0000.}
\doi{10.1109/ACCESS.2017.DOI}

\title{On the Perturbed States for Transformed Input-robust Reinforcement Learning}

\author{{Tung M. Luu$^\ast$}, {Haeyong Kang$^\ast$}, {Tri Ton}, {Thanh Nguyen}, and {Chang D. Yoo}}

\address[1]{School of Electrical Engineering, Korea Advanced Institute of Science and Technology (KAIST), Daejeon 34141, Republic of Korea}

\markboth
{Author \headeretal: Preparation of Papers for IEEE TRANSACTIONS and JOURNALS}
{Author \headeretal: Preparation of Papers for IEEE TRANSACTIONS and JOURNALS}

\corresp{$^\ast$ Equal contribution, Corresponding author: Chang D. Yoo (e-mail: cd\_yoo@kaist.ac.kr).}

\begin{abstract}
Reinforcement Learning (RL) agents demonstrating proficiency in a training environment exhibit vulnerability to adversarial perturbations in input observations during deployment. This underscores the importance of building a robust agent before its real-world deployment. To alleviate the challenging point, prior works focus on developing robust training-based procedures, encompassing efforts to fortify the deep neural network component's robustness or subject the agent to adversarial training against potent attacks. In this work, we propose a novel method referred to as \textit{Transformed Input-robust RL (TIRL)}, which explores another avenue to mitigate the impact of adversaries by employing input transformation-based defenses. Specifically, we introduce two principles for applying transformation-based defenses in learning robust RL agents: \textit{(1) autoencoder-styled denoising} to reconstruct the original state and \textit{(2) bounded transformations (bit-depth reduction and vector quantization (VQ))} to achieve close transformed inputs. The transformations are applied to the state before feeding it into the policy network. Extensive experiments on multiple \mujoco environments demonstrate that input transformation-based defenses, \ie, VQ, defend against several adversaries in the state observations. The official code is available at \url{https://github.com/tunglm2203/tirl}
\end{abstract}


\titlepgskip=-15pt

\maketitle

\section{Introduction}\label{sec:intro}

Modern deep reinforcement learning (RL) agents \cite{mnih2015human,schulman2015trust,silver2016mastering,fujimoto2018addressing,haarnoja2018soft,luu2021hindsight,kostrikov2021offline,luu2022utilizing} typically rely on deep neural networks (DNNs) as powerful function approximations. However, it has been discovered that even a well-trained RL agent can experience significant failures due to small adversarial perturbations in its input during deployment\cite{huang2017adversarial,lin2017tactics,kos2017delving,behzadan2017vulnerability,pattanaik2018robust}. This vulnerability raises concerns about deploying such agents in safety-critical applications, such as autonomous driving \cite{shalev2016safe,sallab2017deep,pan2017virtual,Gray_2019,you2019advanced,liang2022efficient,he2022robust}. Therefore, the development of techniques to help RL agents withstand adversarial attacks on state observations becomes crucial before their deployment in the real world \cite{huang2017adversarial,zhang2020robust,zhang2021robust,gupta2022rsac, wu2022robust,liu2022robustness,korkmaz2023detecting}.

Several approaches have been suggested in the literature to safeguard against adversarial attacks on an agent's state observations. One avenue of research concentrates on improving the robustness of the agent's DNNs component, such as the policy network or Q-value network, by promoting properties like invariance and smoothness through regularization techniques \cite{shen2020deep,zhang2020robust,oikarinen2021robust,yang2022rorl}. These regularizers yield policy outputs that generate similar actions under bounded input perturbations. However, these defenses do not account for the dynamics of the RL environment and can be vulnerable to stronger attacks \cite{zhang2021robust}. An alternative approach involves adversarial training of the RL agent, in which an adversary is introduced to perturb the agent's input as it takes actions within the environment. The episodes collected under these adversarial conditions are then utilized for training, leading to a more robust RL agent. Perturbations in this context can be induced from the policy or value function \cite{kos2017delving,behzadan2017whatever,pattanaik2018robust,liang2022efficient}, or more recently, from another RL-based adversary \cite{zhang2021robust,sun2021strongest}. RL-based adversaries are typically acquired through online learning. Although training with RL-based adversaries can yield robust agents, it often requires additional samples and computations due to the training of the additional agent involved. Furthermore, online attacks may lead to several unsafe behaviors, posing a potential risk to the control system if adversarial training occurs in a physical environment rather than a simulated one.

The approaches above can be seen as robust training-based defenses, wherein the RL agent undergoes optimization to concurrently fulfill the requirements of an RL task and meet the robust performance criteria. Meanwhile, within the realm of image classification, an alternative strategy for countering adversarial attacks is recognized as input transformation-based defenses \cite{dziugaite2016study,meng2017magnet,lu2017no,liao2018defense,guo2018countering,xu2018feature,prakash2018deflecting,samangouei2018defense,gupta2019ciidefence,jin2019ape,salman2020denoised}. These defenses aim to eliminate adversarial perturbations from the input by applying transformations before presenting it to the classifier. The transformational process commonly relies on denoising techniques to purify perturbations \cite{meng2017magnet,liao2018defense,samangouei2018defense,gupta2019ciidefence,jin2019ape,salman2020denoised,nie2022diffusion} or employs image preprocessing techniques to mitigate the impact of adversaries \cite{dziugaite2016study,lu2017no,guo2018countering,xu2018feature,prakash2018deflecting}. Moreover, as this strategy exclusively modifies the input and not the model itself, input transformation-based defenses could benefit RL agents without substantially altering the underlying RL algorithms, serving as a plug-and-play module. Despite their potential to mitigate adversarial attacks, many of these transformations are tailored to image data \cite{dziugaite2016study,guo2018countering,xu2018feature,prakash2018deflecting} and may not easily extend to vector inputs such as low-dimensional states in continuous control systems. This paper explores a new input transformation-based defensive approach, referred to as \textit{\textbf{T}ransformed \textbf{I}nput-robust \textbf{RL} (\textbf{TIRL})}, against adversarial attacks on state observations by applying input transformations suitable for low-dimensional states in continuous control tasks. To validate the effectiveness of the input transformations, various techniques are investigated, including bit-depth reduction, vector quantization, and autoencoder-style denoisers in continuous control tasks. Our findings indicate that these novel defenses exhibit notable resilience against adversarial attacks. Our contributions are summarized as follows:
\begin{itemize}[leftmargin=*]
    \item We propose a new method, \textit{Transformed Input-robust RL (TIRL)} with two principles, \ie, (1) \textit{\textbf{Bounded Transformation}} and (2) \textit{\textbf{Autoencoder-styled Denoising}}, to use input transformations to counter adversarial attacks on state observations in deep RL.
    \item Based on the two proposed principles, we explore the best transformation, \ie, vector quantization (VQ), which is beneficial for the RL agent in continuous control tasks for the first time.
    \item Extensive experiments on continuous control tasks from \mujoco environments demonstrates that these defenses can be surprisingly effective against existing attacks.
\end{itemize}
\section{Related Work} \label{sec:related_word}

\textbf{Adversarial perturbed states.} 
Since the discovery of adversarial examples in image classification \cite{szegedy2013intriguing}, corresponding vulnerabilities in deep reinforcement learning (RL) were first demonstrated in \cite{huang2017adversarial}, \cite{lin2017tactics}. Huang et al. \cite{huang2017adversarial} evaluated the robustness of Deep Q-Network (DQN) agents by employing a weak Fast Gradient Sign Method (FGSM) attacker \cite{goodfellow2014explaining} to initiate attacks at every step. Lin et al. \cite{lin2017tactics} focused on attacking specific steps within trajectories, using a planner to devise perturbations that steer the agent toward a target state. Pattanaik et al. \cite{pattanaik2018robust} introduced a more potent attack strategy that leverages both the policy and the $Q$ function, departing from perturbations based solely on the policy. Zhang et al. \cite{zhang2020robust} later introduced the state-adversarial Markov decision process to formally define adversarial attacks on state observations, showing that optimal attacks can be learned as an RL problem. Building upon this foundation, Zhang et al. \cite{zhang2021robust} and Sun et al. \cite{sun2021strongest} introduced RL-based adversaries for black-box and white-box attacks, respectively. Although these attackers appear powerful, their effectiveness often relies on specific assumptions, such as requiring knowledge of the victim's policy \cite{sun2021strongest}. Consequently, the introduced defenses \cite{zhang2020robust,sun2020stealthy,sun2021strongest} might be vulnerable to a more sophisticated, yet unknown, attacker. In contrast, we focus on developing defense strategies that are agnostic to the attacker's characteristics.

\noindent
\textbf{Adversarial training.} 
The robustness of RL agents can be improved by training with adversarial samples \cite{kos2017delving,behzadan2017vulnerability,huang2017adversarial,mo2022attacking}. In the Atari domain, Kos and Song \cite{kos2017delving} and Behzadan et al. \cite{behzadan2017vulnerability} used weak attacks to generate perturbations for training DQN agents, achieving limited improvement. Alternatively, regularization-based methods have emerged to enhance the robustness of deep neural network components in RL algorithms. Zhang et al. \cite{zhang2020robust} introduced a hinge loss regularizer to control the smoothness of the $Q$ function under bounded perturbations, while Russo and Proutiere \cite{russo2021optimal} proposed Lipschitz regularization. In continuous control tasks, Huang et al. \cite{huang2017adversarial} generated attacks using both the policy and $Q$ function, then used the resulting trajectories for training. However, recent work \cite{zhang2020robust} shows this approach is unreliable for enhancing robustness against new attacks. Smoothness regularization has also been proposed to strengthen policy model robustness in both online \cite{zhang2020robust} and offline RL \cite{yang2022rorl}. Based on the theory of \textit{optimal} attacks from \cite{zhang2020robust}, Zhang et al. \cite{zhang2021robust} introduced a training paradigm that alternates between learning the RL agent and a black-box RL-based adversary to develop robustness. Similarly, Sun et al. \cite{sun2021strongest} applied this approach with a white-box RL-based adversary, yielding a more robust RL agent. Training robust RL agents with additional RL-based adversaries requires extra samples and modifications to the underlying algorithms, making the process more complex and sample-intensive \cite{zhang2021robust,sun2021strongest}. Our approach, however, focuses on modifying the agent's input rather than altering the algorithm. This simplifies training and offers greater flexibility when incorporating other RL algorithms, providing a more practical and effective solution for developing robust RL agents.

\noindent
\textbf{Input Transformation-based Defenses.} 
Studies have used input transformations to mitigate adversarial attacks in image classification due to their simplicity and flexibility \cite{meng2017magnet,lu2017no,liao2018defense,guo2018countering,xu2018feature,prakash2018deflecting,samangouei2018defense,gupta2019ciidefence,jin2019ape,salman2020denoised,nie2022diffusion}. Traditional image processing techniques such as image cropping \cite{guo2018countering} and image rescaling \cite{lu2017no} have been used to reduce the effectiveness of adversarial attacks. Powerful denoising models have also been employed to purify input perturbations. For example, Meng and Chen \cite{meng2017magnet} used an autoencoder-style denoiser to reconstruct denoised images from randomly perturbed ones, while Samangouei et al. used GANs \cite{goodfellow2014generative} for image reconstruction. Nie et al. \cite{nie2022diffusion} applied diffusion models \cite{ho2020denoising} to refine adversarial examples before feeding them to classifiers. However, many of these transformations are specifically designed for image data \cite{dziugaite2016study,guo2018countering,xu2018feature,prakash2018deflecting} and may not easily adapt to vector inputs, such as low-dimensional states in continuous control systems. To alleviate those issues, this paper investigates a novel input transformation-based defense that mitigates the impact of adversarial attacks on state observations for RL agents.
\section{Preliminaries}\label{sec:background}
This section presents the general reinforcement learning framework and an RL algorithm, Soft Actor-Critic (SAC), before introducing our novel input-transformation-based defenses. To evaluate the effectiveness of our method with SAC, we also describe test-time adversarial attacks.

\subsection{Reinforcement Learning}
A reinforcement learning (RL) environment is modeled by a Markov decision process (MDP), defined as a tuple of $(\mathcal{S}, \mathcal{A}, p, r, \gamma)$, where $\mathcal{S}$ is the state space, $\mathcal{A}$ is the action space, $p=Pr(s_{t+1}|s_t, a_t)$ is the transition probability distribution which is usually unknown, $r: \mathcal{S} \times \mathcal{A} \times \mathcal{S} \rightarrow \mathbb{R}$ is the reward function, and $\gamma \in [0,1)$ is a discount factor. An agent executes actions based on a policy $\pi: \mathcal{S}\rightarrow\mathcal{A}$. In this work, we use the terms \textit{state} and \textit{state observation} to indicate the underlying state of the environment and the state perceived by the agent, respectively. If there are no attacks, both of them are identical. The target of RL agent is to maximize the expected discounted return $\mathbb{E}_{\pi,p}\left[\sum_{t=0}^{\infty}\gamma^t r(s_t,a_t,s_{t+1})\right]$, which is the expected cumulative sum of rewards when following the policy $\pi$ in the MDP. The state value function can measure this objective as follows:
\begin{equation*}
  V^{\pi}(s):=\mathbb{E}_{\pi,p}\left[\sum_{t=0}^{\infty}\gamma^t r(s_t,a_t,s_{t+1}) | s_0=s\right],
\end{equation*}
or the state-action value function:
\begin{equation*}
    Q^{\pi}(s,a) := \mathbb{E}_{\pi, p}\left[\sum_{t=0}^{\infty}\gamma^tr(s_t,a_t,s_{t+1})|s_0=s, a_0=a\right].
\end{equation*}

\subsection{Training Soft Actor-Critic} 
Soft Actor-Critic (SAC) \cite{haarnoja2018soft} is an actor-critic off-policy RL algorithm widely used for solving continuous control tasks. This paper adopts SAC as the backbone of the proposed method for robustness evaluation due to its stability and sample efficiency in continuous control benchmarks. However, our method can also be easily extended to other algorithms, such as Proximal Policy Optimization (PPO) \cite{schulman2017proximal} and Twin Delayed Deep Deterministic Policy Gradient (TD3) \cite{fujimoto2018addressing}. 

SAC alternates between the soft policy evaluation and soft policy improvement steps to learn the optimal stochastic policy that maximizes the expected returns and entropy of actions. Denoting $\pi_{\theta}$ as a stochastic policy parameterized by $\theta$. In soft policy evaluation step, SAC learns two critic networks, denoted as $Q_{\phi_1}$ and $Q_{\phi_2}$, parameterized by $\phi_1$ and $\phi_2$, respectively, by minimizing the soft Bellman residual:
\begin{equation}\label{eq:q_loss}
    J_Q(\phi_i) = \mathbb{E}_{(s, a, s', r)\sim \mathcal{R}, a'\sim\pi_{\theta}} \left[(Q_{\phi_i}(s, a) - y)^2\right],
\end{equation}
where, $y = r + \gamma[\min_{i=1,2}Q_{\hat{\phi}_i}(s',a') - \alpha\log \pi_{\theta}(a'|s')]$, $\mathcal{R}$ is the replay buffer, $\alpha$ is the temperature parameter. The parameter $\hat{\phi}_i$ denotes the exponential moving average (EMA) of $\phi_i$, which empirically improves training stability in off-policy RL algorithms \cite{fujimoto2018addressing,haarnoja2018soft}. In soft policy improvement step, the actor policy $\pi_{\theta}$ is optimized by maximizing the following objective:
\begin{equation}\label{eq:pi_loss}
    J_{\pi}(\theta) = \mathbb{E}_{s\sim\mathcal{R}, a\sim\pi_{\theta}}\left[\min_{i=1,2}Q_{\phi_i}(s, a) - \alpha \log (\pi_{\theta}(a|s))\right],
\end{equation}
where, the stochastic policy $\pi_{\theta}(a|s)$ is a parametric tanh-Gaussian that given the state $s$, an action $a$ is sampled from $ \tanh{(\mu(s) + \sigma(s)\epsilon)}$, with $\epsilon \sim \mathcal{N}(0, 1)$, and $\mu$ and $\sigma$ are parametric mean and standard deviation. Finally, the temperature $\alpha$ is learned with following objective:
\begin{equation}
    J(\alpha) = \mathbb{E}_{s \sim\mathcal{R}, a \sim \pi_{\theta}}\left[-\alpha\log\pi_{\theta}(a|s)-\alpha\bar{\mathcal{H}}\right],
\end{equation}
where, $\bar{\mathcal{H}}\in \mathbb{R}$ is the target entropy hyperparameter that policy tries to match, which is usually set to $-|\mathcal{A}|$ in practice with $\mathcal{A}$ is the action space.

\subsection{Test-time Adversarial Attacks} \label{subsec:test_time_attack}

\begin{figure}[t]
    \centering
    \includegraphics[width=0.47\textwidth]{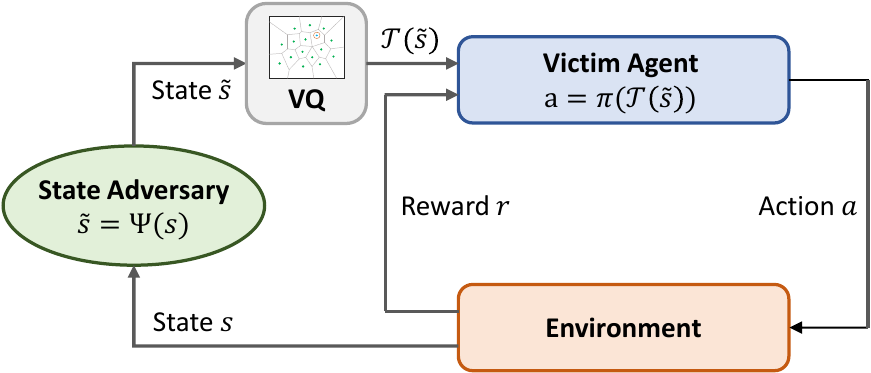}
    \caption{\textit{\textbf{Transformed Input-robust RL (TIRL)-Vector Quantization (VQ)}}: Reinforcement learning with the state adversary at test time. The state $s$ is adversarially perturbed by the adversary $\Psi(s)$ into $\tilde{s}$, which is then transformed by the transformation $\mathcal{T}$ before being fed to the agent.}
    \label{fig:sa_mdp}
    \vspace{-0.16in}
\end{figure}

After training, the agent is deployed within its designated environment and operates according to a fixed pretrained policy $\pi_{\theta}$, \ie, the parameter $\theta$ is frozen. During this testing phase, an adversary can observe interactions between the victim agent and the environment, including states, actions, and rewards, to mislead the agent. However, the adversary does not possess knowledge of the environment's dynamics nor the capability to alter the environment directly. We focus on a typical \textit{state adversary} that targets the manipulation of the state observation of the RL agent \cite{huang2017adversarial, lin2017tactics, kos2017delving, pattanaik2018robust, weng2019toward, fischer2019online, shen2020deep, russo2021optimal, zhang2021robust, sun2021strongest}. The MDP with a state adversary involved is formally defined by the state-adversarial Markov decision process (SA-MDP) introduced in \cite{zhang2020robust}. Let $\Psi: \mathcal{S} \rightarrow \mathcal{S}$ represent a state adversary. At each time step, the adversary $\Psi$, equipped with a certain budget $\epsilon$, is allowed to inject perturbations to state $s$ returned by the environment before the agent perceives them, as illustrated in Fig. \ref{fig:sa_mdp}. It is important to note that the underlying states of the environment, returned rewards, and executed actions remain unaffected. This setting aligns with many realistic scenarios, such as sensor measurement errors, noise in sensory signals, or man-in-the-middle (MITM) attacks on a deep RL system. For instance, in robotic manipulation, an attacker could inject imperceptible noises into the camera, capturing an object without altering the actual object's location. As in many of the existing works investigating adversarial robustness of deep RL \cite{huang2017adversarial, pattanaik2018robust, zhang2020robust, oikarinen2021robust, zhang2021robust, sun2021strongest}, we focus on the $\ell_{\infty}$-norm attack, in which the adversary $\Psi$ is constrained to perturb the state $s\in\mathcal{S}$ into another ``neighboring'' state $\tilde{s}$ within an $\epsilon$-radius $\ell_{\infty}$-norm ball around $s$. Formally, the perturbed state $\tilde{s} = \Psi(s) \in \mathcal{B}_{\epsilon}(s)$, with $ \mathcal{B}_{\epsilon}(s) = \{\tilde{s}: {\|s - \tilde{s}\|}_{\infty} \leq \epsilon\}$. The details of state adversaries are described in Section \ref{subsec:experimental_setup}.
\section{Transformed Input-robust RL (TIRL)}\label{sec:method}
In this section, we theoretically analyze the effectiveness of input transformation-based defenses in improving the robustness of RL agents. Subsequently, two principles are proposed to design transformation defenses for RL agents: 1) \textit{\textbf{Bounded Transformation}} and 2) \textit{\textbf{Autoencoder-styled Denoising}}. Based on these principles, we explore potential transformations to counter adversarial attacks.

\subsection{Effective Input Transformation Defenses}
We first characterize the agent's performance changes before and after attacks by measuring the difference in value functions between the policy in MDP and SA-MDP. Let $\pi$ be the policy, $V^{\pi}(s)$ be its state value function in the regular MDP, and $V^{\pi\circ\Psi}(s)$ be its state value function in the SA-MDP with the adversary $\Psi$. Prior work \cite{zhang2020robust} provides an upper bound on the difference between
value functions caused by an optimal adversary. Specifically, theorem 5 in \cite{zhang2020robust} states that, under the optimal adversary $\Psi^*$ in SA-MDP, the performance gap between $V^{\pi}(s)$ and $V^{\pi\circ\Psi^*}(s)$ can be bounded as follows:
\begin{equation}\label{eq:samdp_bound}
    \small
    \max_{s\in\mathcal{S}}\{V^{\pi}(s)-V^{\pi\circ\Psi^*}(s)\} \leq \kappa \max_{s\in\mathcal{S}}\max_{\tilde{s}\in \mathcal{B}_{\epsilon}(s)}D_{TV}(\pi(\cdot|s), \pi(\cdot|\tilde{s}))
\end{equation}
where, $D_{TV}(\pi(\cdot|s), \pi(\cdot|\tilde{s}))$ represents the total variance distance between $\pi(\cdot|s)$ and $\pi(\cdot|\tilde{s})$, $\kappa$ is a constant independent of $\pi$, $\mathcal{B}_{\epsilon}(s)$ is defined as in Section \ref{subsec:test_time_attack}, and $\pi\circ\Psi^*$ denotes the policy under perturbations, \ie , $\pi(a|\Psi^*(s))$. The detailed proof for Eq. \ref{eq:samdp_bound}, which is based on tools developed in constrained policy optimization \cite{achiam2017constrained}, can be found in the Appendix of \cite{zhang2020robust}.

Based on the upper bound between value functions, \cite{zhang2020robust,shen2020deep} proposed regularizing $D_{TV}(\pi(\cdot|s), \pi(\cdot|\tilde{s}))$ during policy training. In contrast, we focus on simple yet effective input transformations to reduce the performance gap. Let $\mathcal{T}$ be a transformation that maps $\mathcal{S} \rightarrow {\bar{S}}$, where ${\bar{S}}$ is a set of all transformed states. This transformation can be applied during the training and testing phases or exclusively during the testing phase. Our following proposition establishes the connection between the performance gap and the input transformations.

\prop{
\textit{Consider a $K$-Lipschitz continuous policy $\pi$ parameterized by the Gaussian distribution with a constant variance independent of state for a regular MDP.
Let the corresponding value function be $V^{\pi}(s)$. Define $\mathcal{T}_{t}$ and $\mathcal{T}_{d}$ as the transformations applied during training and testing, respectively. Under the optimal adversary $\Psi^*$ in SA-MDP, for all $s\in\mathcal{S}$ we have:
\begin{equation}\label{eq:performance_bound}
	\small
	\max_{s\in\mathcal{S}}\{V^{\pi \circ \mathcal{T}_t}(s)-V^{\pi\circ \mathcal{T}_d \circ \Psi^*}(s)\} \leq \zeta \max_{s\in\mathcal{S}}\max_{\tilde{s}\in \mathcal{B}_{\epsilon}(s)} {\|\mathcal{T}_t(s)-\mathcal{T}_d(\tilde{s})\|}_2,
\end{equation}
where, $\zeta$ is a constant independent of $\pi$. Here, $\pi\circ\mathcal{T}_t $ and $\pi\circ\mathcal{T}_d \circ \Psi^*$ denote $\pi(\cdot|\mathcal{T}_t(s))$ and $\pi(\cdot|\mathcal{T}_d(\Psi(s)))$, respectively.}
}\label{prop:performance_bound} \\

\noindent
\underline{\textit{Proof:}}
Based on Pinsker’s inequality, we first upper bound the total variance distance in RHS of Eq. (\ref{eq:samdp_bound}) by Kullback-Leibler (KL) divergence:
\begin{equation}\label{eq:tv_kl}
    D_{TV}(\pi(\cdot|s), \pi(\cdot|\tilde{s})) \leq \sqrt{\frac{1}{2}D_{KL}\left(\pi(\cdot|s) \parallel \pi(\cdot|\tilde{s})\right)}.
\end{equation}
Assuming the state space has dimension $d$, and given that the policy is Gaussian with constant independence variance, we denote $\pi(\cdot|s) \sim \mathcal{N}(\mu_s,\Sigma_s)$ and $\pi(\cdot|\tilde{s}) \sim \mathcal{N}(\mu_{\tilde{s}},\Sigma_{\tilde{s}})$, where $\mu_s, \mu_{\tilde{s}} \in \mathbb{R}^d$ are produced by neural networks $\mu_{\theta}(s), \mu_{\theta}(\tilde{s}) $, repectively, and $\Sigma$ is a diagonal matrix independent of the state, i.e., $\Sigma_s = \Sigma_{\tilde{s}} = \Sigma$. Using the $K$-Lipschitz assumption of the policy network, we can bound the KL divergence as follows:
\begin{equation}\label{eq:kl_lipschitz}
	\begin{aligned}
		D_{KL}\left(\pi(\cdot|s) \parallel \pi(\cdot|\tilde{s})\right) &= \frac{1}{2}(\log\frac{|\Sigma_{\tilde{s}}|}{|\Sigma_s|} - d + tr(\Sigma_{\tilde{s}}^{-1}\Sigma_s) \\
        & \quad \quad + (\mu_{\tilde{s}} - \mu_{s})^{\top}\Sigma_{\tilde{s}}^{-1}(\mu_{\tilde{s}} - \mu_{s})) \\
		&\leq L{\|\mu_{\tilde{s}} - \mu_{s}\|}^2_2 \\
		&= L{\|\mu_{\theta}(s) - \mu_{\theta}(\tilde{s})\|}_2^2\\
		&\leq LK{\|s - \tilde{s}\|}_2^2.
	\end{aligned}
\end{equation}
The first inequality holds because $\Sigma_{\tilde{s}}, \Sigma_{s}$ are positive, then ensuring there exists a positive constant $L\in\mathbb{R}^+$ that satisfies this inequality. The second inequality holds because the policy network is $K$-Lipschitz continuous. Let $\mathcal{T}_t$ and $\mathcal{T}_d$ be the transformations applied into state $s$ and $\tilde{s}$, respectively. By combining Equation (\ref{eq:samdp_bound}), (\ref{eq:tv_kl}), and (\ref{eq:kl_lipschitz}), we obtain:
\begin{equation}
	\text{\textit{LHS} of (\ref{eq:samdp_bound})}\leq \zeta\max_{s\in\mathcal{S}}\max_{\tilde{s}\in \mathcal{B}_{\epsilon}(s)}{\|\mathcal{T}_t(s) - \mathcal{T}_d(\tilde{s})\|}_2
\end{equation}
where $\zeta:=\frac{1}{\sqrt{2}}\kappa\sqrt{LK}$. The proof is complete. $\hfill \square$ 

\subsection{Input Transformation Principles}
\textit{\textbf{Proposition}}~\ref{prop:performance_bound} says that the performance gap can be bounded by the difference between the ``origin'' state and ''perturbed'' state in the transformed space, \ie, ${\|\mathcal{T}_t(s)-\mathcal{T}_d(\tilde{s})\|}_2$. It suggests two distinct principles for reducing the performance gap through transformations. We summarize our two principles as follows: 
\begin{enumerate}[leftmargin=*]
    \item \textit{\textbf{Bounded Transformation}}: If $\mathcal{T}_t$ and $\mathcal{T}_d$ are used for both training and testing, we can design $\mathcal{T}_t$ and $\mathcal{T}_d$ such that, after transformation, the difference between the state and its perturbed version is small, \ie, $\max_{\tilde{s}\in \mathcal{B}_{\epsilon}(s)}{\|\mathcal{T}_t(s) - \mathcal{T}_d(\tilde{s})\|}_2$ is minimized.

    \item \textit{\textbf{Autoencoder-styled denoising}}: If $\mathcal{T}_{t}$ is the identity function, \ie, $\mathcal{T}_t$ is not used during training, we need to design $\mathcal{T}_d$ such that it minimizes $\max_{\tilde{s}\in\mathcal{B}_{\epsilon}(s)}{\|s - \mathcal{T}_d(\tilde{s})\|}_2$. In other words, $\mathcal{T}_d$ needs to reconstruct origin state $s$ from the perturbed state $\tilde{s}$. This naturally leads to the selection of denoisers for $\mathcal{T}_d$.
\end{enumerate} 

\noindent
In the autoencoder-styled denoising approach, we can theoretically maintain the performance under attacks, $V^{\pi\circ\mathcal{T}_d\circ\Psi^*}(s)$, close to the natural performance $V^{\pi}(s)$ as long as the denoiser can effectively reconstruct the original state $s$ from the perturbed state $\tilde{s}$. However, this approach might be vulnerable to fully white-box gradient-based attacks, where adversaries have access to the entire policy's architecture, including its parameters and the denoiser. In the bounded transformation approach, the strength of these defenses lies in their non-differential nature, making them more resilient to white-box attacks. However, due to the use of transformations during training, it is not guaranteed that the natural performance under the transformation, $V^{\pi\circ\mathcal{T}_t}(s)$, is identical to $V^{\pi}(s)$, as some information in the state may be lost. Nevertheless, as long as $\mathcal{T}_{t}$ retains the essential information from the original space, we can expect $V^{\pi\circ\mathcal{T}_t}(s)$ to be close to $V^{\pi}(s)$, as empirically observed in our experiments. We present input transformation-based defenses following these two principles in the following parts. 

\begin{figure}[t]
    \centering
    \includegraphics[width=0.4\textwidth]{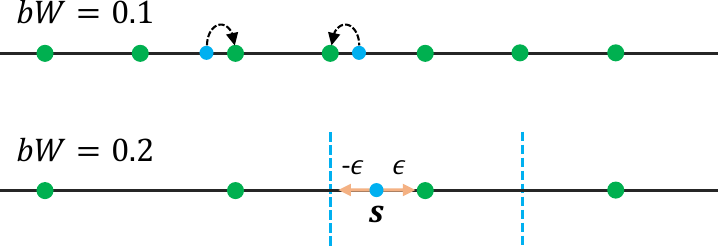}
    \caption{Top: Illustration of using bit-depth reduction to quantize the 1-D state space. The \textcolor{Cerulean}{state} is assigned to the \textcolor{ForestGreen}{closest point}. Bottom: The current state and its perturbed versions may be assigned the same value if $\epsilon$ is not too large. A larger bin width ($bW$) led to greater robustness.}
    \label{fig:bdr}
    \vspace{-0.16in}
\end{figure}

\subsection{Various Input Transformations}
We investigate various input transformations to satisfy the two principles. In each principle, we can summarize as follows: (1) \textit{Bounded Transformation (bit-depth reduction and vector quantization (VQ)} to achieve close transformations of origin and perturbed state and (2) \textit{Autoencoder-styled Denoising, i.e., VAED} to reconstruct origin state.

\subsubsection{Bounded Transformation: Bit-Depth Reduction}
 The bit-depth reduction (BDR) transformation, initially explored as a defense in image classification tasks \cite{xu2018feature,guo2018countering}, is adopted here for our control tasks. The BDR performs uniform quantization for each input dimension, assigning each value to the nearest quantized point (Fig. \ref{fig:bdr} top). If the adversary's budget $\epsilon$ is not too large, the state $s$ and its perturbed state $\tilde{s}$ within the $\epsilon$-radius can be assigned the same value after transformation (Fig. \ref{fig:bdr} bottom). The BDR adheres to the bounded transformation principle, keeping ${\|\mathcal{T}_t(s) - \mathcal{T}_d(\tilde{s})\|}_2$ small overall. The original design of BDR for images does not directly apply to continuous vector states due to varying value ranges and the continuous nature of the data. To address the limited flexibility, we redesigned BDR by partitioning each dimension into equally spaced bins, controlled by the bin width ($bW$) parameter, and rounding real-valued scalars to the nearest bin value. To balance robustness and natural performance, we adjust the $bW$ parameter: increasing the bin width enhances defense against attacks but may reduce overall performance and vice versa.

\begin{figure}[t]
    \centering
    \includegraphics[width=0.3\textwidth]{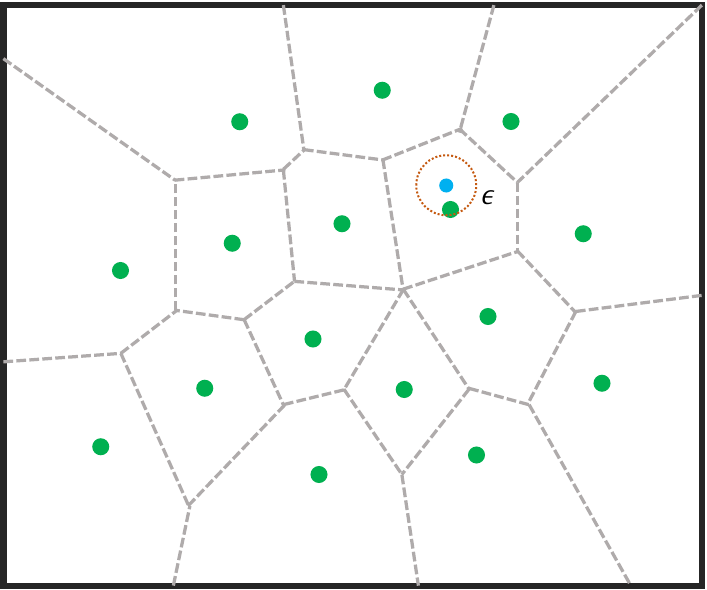}
    \caption{Illustration of using vector quantization to reduce the space of adversarial attacks in a 2-D state space. The green dots represent the centroids of clusters, while the gray dotted lines mark the boundaries between clusters. Each \textcolor{Cerulean}{state} is assigned to closest \textcolor{ForestGreen}{centroid}. Fewer clusters result in a sparser state space, leading to greater robustness.}
    \label{fig:vanilla_vq}
    \vspace{-0.16in}
\end{figure}

\subsubsection{Bounded Transformation: Vector Quantization}
We propose a new sparse input transformation using vector quantization (VQ) to discretize the entire state space into discrete vectors more strongly, as shown in Fig. \ref{fig:vanilla_vq}. To perform VQ for state space, we employ the K-means algorithm. Specifically, given a set of state vectors $\{{s}_i\}_{i=1}^N$, the K-means algorithm partitions the $N$ points into $K$ clusters. Subsequently, each state is then represented by its nearest centroid in the codebook ${C}=[{c}_1, {c}_2, \dots, {c}_K]$. The training of K-means is achieved by minimizing the following objective:
\begin{equation}\label{eq:vq_learning}
    \min_{{C}, \{{p}_i\}} \sum_{i=1}^N{\|{s}_i - {C}{p}_i\|}_2^2
\end{equation}
where ${p}_i\in \{0,1\}^K$ is a one-hot vector with the value 1 indicating the index of the centroid nearest to data point $s_i$ and $K$ controls the codebook size. In the backward pass, the VQ module is treated as an identity function, referred to as straight-through gradient estimation~\cite{bengio2013estimating}. The above objective is also convenient for learning the codebook online, which is suitable for concurrent training with RL algorithms since the data collected by the RL agent continuously evolves during training time. Unlike BDR, after transformation, VQ can produce a more ``sparse'' state space, thus significantly reducing the space of attacks, leading to the agent being more robust against adversarial attacks. 

\subsubsection{Autoencoder-styled Denoising}
Training denoisers requires perturbed states, which are additionally collected by interacting with the environment under attack, making the process inefficient. To obtain stable denoising results, we only apply the autoencoder-styled denoiser \cite{meng2017magnet, liao2018defense} after completing the training of the RL agent. Moreover, sampling data under online adversarial attacks can result in potentially unsafe behaviors, especially in a physical system. For instance, in manipulation tasks using a robotic arm, collecting samples under attacks might lead to the breakage of objects. To mitigate these challenges, we leverage samples from the replay buffer collected during the training of RL agents for denoiser training. To generate perturbed states, we use \textit{Action Diff}~\cite{zhang2020robust} and \textit{Min Q}~\cite{pattanaik2018robust} attacks, as they only require a policy or $Q$-value network and do not necessitate interaction with the environments. This strategy helps reduce costs for acquiring additional data and avoids unsafe behaviors. In this approach, we must select an appropriate $\epsilon$-radius for generating perturbations. If the $\epsilon$ is too large, the denoiser might be unable to reconstruct the original states.

\newcommand{\MODIFY}[1]{\textcolor{blue}{#1}}
\begin{algorithm}[h]
    \small 
    \caption{\MODIFY{TIRL} for training the agent with SAC}
    \label{alg:tirl}
    \begin{algorithmic}[1]
        \STATE \textbf{Input}: $T$ training steps, initialize policy network $\pi_{\theta}$ and critic networks $Q_{\phi_1}$, $Q_{\phi_2}$ with parameters $\theta$, $\phi_1$, and $\phi_2$, the target critic networks $\hat{\phi_1} \leftarrow \phi_1, \hat{\phi_2} \leftarrow \phi_2$, initialize replay buffer $\mathcal{R}$. Transformation $\MODIFY{\mathcal{T}}$ and its hyperparameter, \ie, $bW$ for BDR or $K$ for VQ.
        \vspace{0.05in}
        
        \FOR {$ t = 0 \text{ \textbf{to} } T $}
        \STATE Execute action $a_i \sim \pi_{\theta}(\MODIFY{\mathcal{T}}(s_i))$ into environment.
        \STATE Store transition $\{s_t, a_t, r_t, s_{t+1}\}$ into $\mathcal{R}$.
        \STATE Sample a mini-batch of $N$ samples $\{s_j, a_j, r_j, s'_{j}\}$ from $\mathcal{R}$.
        \STATE $y_j = r_j + \gamma[\min_{i=1,2}Q_{\hat{\phi}_i}(s'_j,a') - \alpha\log \pi_{\theta}(a'|\MODIFY{\mathcal{T}}(s'_j))]$, for all $j\in [N]$, and $a'\sim \pi_{\theta}(\cdot|\MODIFY{\mathcal{T}}(s'_j))$.
        \STATE Train critic:
        \begin{center}
            $J(\phi_i) = \frac{1}{N}\sum_j (Q_{\phi_i}(s_j, a_j) - y_j)^2$
        \end{center}
        
        \STATE Train policy: 
        \begin{center}
            $J(\theta) = \frac{1}{N}\sum_j\min_{i=1,2}Q_{\phi_i}(s_j, a) - \alpha \log (\pi_{\theta}(a|\MODIFY{\mathcal{T}}(s_j)))$
        \end{center}
        where, $a\sim \pi_{\theta}(\cdot|\MODIFY{\mathcal{T}}(s_j))$.
        \IF {TIRL-VQ is used}
        \STATE Update codebook $C$ for TIRL-VQ using Eq. (\ref{eq:vq_learning}).
        \ENDIF
        \ENDFOR
    \end{algorithmic}
\end{algorithm}
\begin{algorithm}[h]
    \small 
    \caption{\MODIFY{TIRL} for the agent at test time}
    \label{alg:tirl_test}
    \begin{algorithmic}[1]
        \STATE \textbf{Input}: $N$ episodes for evaluation, pretrained policy network $\pi_{\theta}$ from Alg. \ref{alg:tirl}. Adversary $\Psi$ and $\epsilon$-radius. Transformation $\MODIFY{\mathcal{T}}$. 
        \vspace{0.05in}
        
        \FOR {$ i = 0 \text{ \textbf{to} } N $}
        \WHILE{not done}
        \STATE Get state $s$ from environment
        \STATE Get perturbed state $\tilde{s} = \Psi(s)$
        \STATE Execute action $a \sim \pi_{\theta}(\MODIFY{\mathcal{T}}(\tilde{s}))$ into environment.
        \ENDWHILE
        \ENDFOR
    \end{algorithmic}
\end{algorithm}

\noindent
The pseudo-code for applying BDR or VQ during agent training is shown in Algorithm \ref{alg:tirl}. Note that we only apply these transformations to the input of the policy network. After training is complete, the transformations are applied to mitigate adversarial attacks, as demonstrated in Algorithm \ref{alg:tirl_test}.

\section{Experiments}\label{sec:experimental}
In this section, we evaluate the efficacy of our defenses across continuous control tasks. Section~\ref{subsec:experimental_setup} details the setup and adversaries used to evaluate the robustness. The experiment in Section~\ref{subsec:graybox} evaluates performance in a gray-box setting, where adversaries can access the model architecture and parameters but are unaware of the applied transformations. The experiment in Section~\ref{subsec:whitebox} focuses on a white-box setting, where the adversaries are aware of the defense strategies.

\subsection{Experimental Setup}\label{subsec:experimental_setup}

\begin{figure*}[t]
\centering
\begin{tabular}{cccccc}
\includegraphics[width=0.165\textwidth]{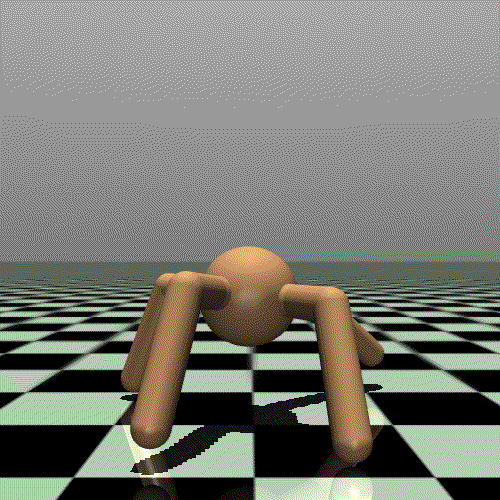} &
\includegraphics[width=0.165\textwidth]{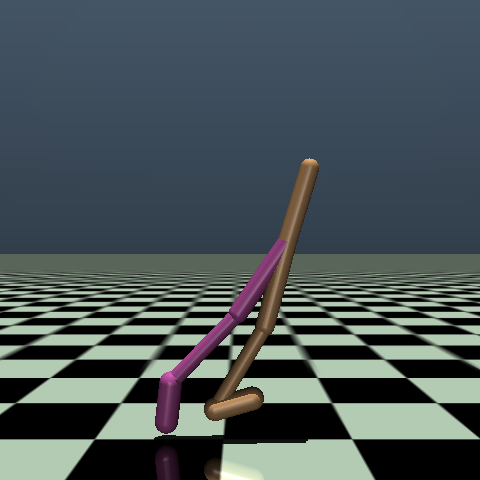} &
\includegraphics[width=0.165\textwidth]{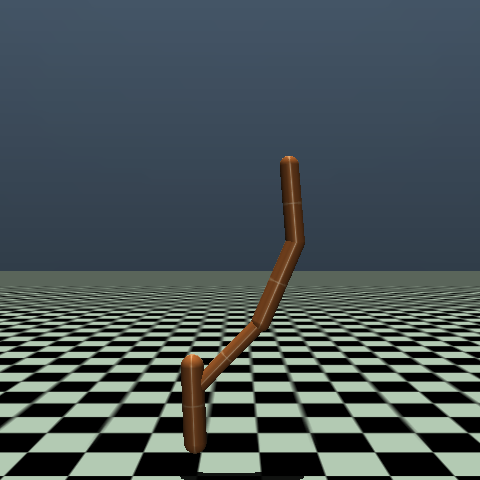} &
\includegraphics[width=0.165\textwidth]{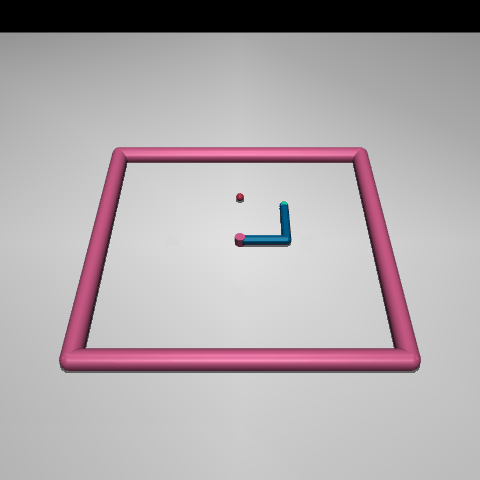} &
\includegraphics[width=0.165\textwidth]{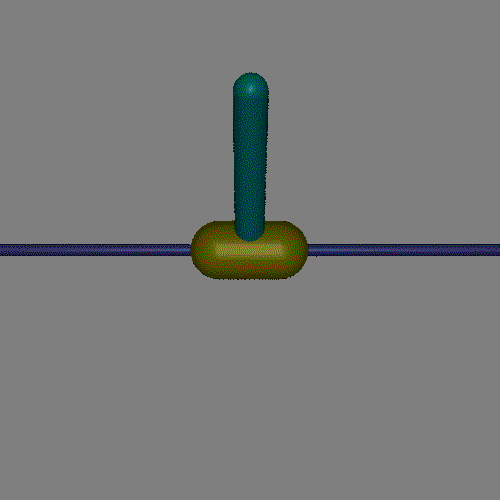} 
\end{tabular}

\caption{The five \mujoco environments in OpenAI Gym \cite{brockman2016openai} are used to evaluate the robustness for SAC-based TILR.}
\label{fig:mujoco_environments}
\vspace{-0.1in}
\end{figure*}

\textbf{Environments and Baseline.} We evaluate the effectiveness defenses on five \mujoco environments, as commonly used in the literature \cite{lillicrap2015continuous, fujimoto2018addressing, zhang2020robust}, including Walker2d, Hopper, Ant, Inverted Pendulum, and Reacher, as shown in Fig. \ref{fig:mujoco_environments}. We use SAC as a base RL algorithm with the implementation from \cite{d3rlpy}, the hyperparameters for the five environments similar to those in \cite{haarnoja2018soft}. When applying the input transformations, we maintain the same hyperparameters for the underlying RL algorithm and only conduct a grid search for the transformation's hyperparameter. We run Walker2d and Ant $3 \times 10^6$ steps, Hopper $10^6$ steps, and Inverted Pendulum and Reacher $2\times 10^5$ steps. 

\noindent
\textbf{Adversaries.} We attack the trained policies with multiple existing attack methods, as used in \cite{huang2017adversarial, zhang2020robust, zhang2021robust, sun2021strongest}, including \textit{Random}, \textit{Action Diff}, \textit{Min Q}, \textit{Robust Sarsa (RS)}, and \textit{Policy Adversarial Actor Director (PA-AD)}. Specifically, given an attack budget $\epsilon$, a state $s$ is adversarially perturbed into state $\tilde{s}$ by adversaries as follows:
\begin{itemize}[leftmargin=*]
    \item \textit{Random} uniformly samples perturbed states within an $\epsilon$-radius $\ell_{\infty}$-norm ball.
    \item \textit{Action Diff} \cite{zhang2020robust} is an attack method relied on the agent's policy. It directly finds the perturbed states within an $\epsilon$-radius $\ell_{\infty}$-norm ball to satisfy: $\max_{\tilde{s}\in\mathcal{B}_{\epsilon}(s)}D_{KL}(\pi_{\theta}(\cdot|s)||\pi_{\theta}(\cdot|\tilde{s}))$ with $D_{KL}$ is the Kullback-Leibler divergence.
    \item \textit{Min Q} \cite{pattanaik2018robust} relies on the agent's policy and $Q$ function to perform attacks. It generates the perturbed states within an $\epsilon$-radius $\ell_{\infty}$-norm ball to satisfy: $\min_{\tilde{s}\in\mathcal{B}_{\epsilon}(s)}Q_{\phi}(s, \pi_{\theta}(\tilde{s}))$
    \item \textit{Robust Sarsa} (RS) \cite{zhang2020robust} is performed similar to \textit{Min Q} attack with a robust action-value function.
    \item \textit{Policy Adversarial Actor Director} (PA-AD) \cite{sun2021strongest} is currently the most powerful attack on RL agents, as it learns the optimal adversary using RL. Specifically, given a pretrained policy $\pi_{\theta}$, PA-AD learns an adversarial policy $\pi_{adv}$, which takes the state $s$ as input to generate the adversarial state $\tilde{s}$. The adversarial policy is trained using Proximal Policy Optimization (PPO) \cite{schulman2017proximal}.
\end{itemize}

\noindent
Generally, the strength of attackers is ranked in ascending order as follows: \textit{Random} < \textit{Action Diff} < \textit{Min Q} < \textit{RS} < \textit{PA-AD}. For \textit{Action Diff}, \textit{Min Q}, and \textit{RS}, we use projected gradient descent (PGD) to generate the optimal perturbed state, as in \cite{kurakin2016adversarial,lin2017tactics,pattanaik2018robust,zhang2020robust}. For example, the perturbed state $\tilde{s}$ is obtained by PGD in \textit{Action Diff} attack as follows:
\begin{equation}\label{eq:action_diff}
    \tilde{s}^{k+1} = \tilde{s}^{k} + \eta \; \text{proj} [\nabla_{\tilde{s}^k} D_{KL}(\pi_{\theta}(\cdot|s)||\pi_{\theta}(\cdot|\tilde{s}^{k}))] 
\end{equation}
where $\tilde{s}^0 = s$, $k = 0, \dots, H - 1$, $\text{proj}[.]$ is a projection to $\mathcal{B}_{\epsilon}(s)$, $\eta$ is the step size, and $H$ is the number of iterations. Through experiments, we use 10-step PGD for optimization for these attacks (\ie, $H=10$) with a step size $\eta = 0.1$. For PA-AD, we follow the setup in \cite{sun2021strongest} to evaluate pre-trained SAC policies. The architecture of the adversarial policy is similar to the SAC policy, consisting of two 256-dimensional hidden layers followed by \texttt{ReLU} activations, except for the last layer, which uses \texttt{Tanh} activation. The $\epsilon$-radius of attackers for each environment is shown in the tables. \\

\begin{table}[]
\renewcommand{\arraystretch}{1.3}
\centering
\caption{Main hyperparameter of TIRL.}
\begin{tabular}{|c@{\hspace{3pt}}|c@{\hspace{3pt}}|c@{\hspace{3pt}}|c@{\hspace{3pt}}|c@{\hspace{3pt}}|c@{\hspace{3pt}}|}
\hline
\textbf{TIRL}/ \textbf{Env.}  & \textbf{Ant}            & \textbf{Hopper}         & \textbf{Walker2d} & \textbf{Pendulum}       & \textbf{Reacher}        \\ \hline
BDR ($bW$)           & 0.05           & 0.15           & 0.1      & 0.05           & 0.4            \\ \hline
VQ ($K$)             & $5\times 10^5$ & $5\times 10^5$ & $10^5$   & $3\times 10^5$ & $2\times 10^5$ \\ \hline
AED ($\epsilon$)     & 0.15           & 0.1            & 0.05     & 0.3            & 0.25           \\ \hline
VAED ($\epsilon$)    & 0.15           & 0.1            & 0.05     & 0.3            & 0.25           \\ \hline
\end{tabular}
\label{tab:main_hyperparam}
\vspace{-0.15in}
\end{table}
\begin{table*}[t]
\renewcommand{\arraystretch}{1.3}
\centering
\caption{Average episode return $\pm$ standard deviation of SAC agents and SAC-TIRL with various transformation defenses in the gray-box setting. Results are averaged over 5 runs, with evaluations conducted over 50 episodes per run. Following \cite{kostrikov2021offline}, we \textbf{bold} all scores within 5 percent of the maximum in each attack and environment ($\geq$ 0.95 $\cdot$ max).}
\label{tab:gray_box}
\begin{tabular}{l@{\hspace{4pt}}|l@{\hspace{4pt}}|c@{\hspace{6pt}}|c@{\hspace{6pt}}c@{\hspace{6pt}}c@{\hspace{6pt}}c@{\hspace{6pt}}c@{\hspace{6pt}}c@{\hspace{6pt}}}
    \noalign{\hrule height 0.8pt}
    \textbf{Environment}   & \textbf{Method}  &   \textbf{Natural Returns} & \textbf{Random} & \textbf{Action Diff}   & \textbf{Min Q}   &  \textbf{RS} & \textbf{PA-AD} & \textbf{Average}\\
    \noalign{\hrule height 0.8pt}
    \multirow{5}{*}{\specialcell{\textbf{Ant}\\(state-dim: 111)\\$\epsilon$: 0.15}}   
    & SAC   
    & \textbf{6859  $\pm$ 126}   &  \textbf{5637 $\pm$ 293}   & 454  $\pm$ 335    &  370 $\pm$ 279      & 463 $\pm$ 216       &  -962 $\pm$ 341  & 2136.8  \\   
    & SAC-TIRL-BDR  
    & \textbf{6543  $\pm$ 506}   &  \textbf{5445 $\pm$ 477}   & 1113 $\pm$ 310    &  1071 $\pm$ 564     & 910 $\pm$ 329       &  899  $\pm$ 367  & 2663.5  \\
    & \textbf{SAC-TIRL-VQ}   
    & 4785  $\pm$ 164   &  4737 $\pm$ 205   & \textbf{4667 $\pm$ 405}    &  \textbf{4595 $\pm$ 90}      & \textbf{4485 $\pm$ 498}      & \textbf{4569$ \pm$ 237}   & \textbf{4639.7}  \\
    & SAC-TIRL-AED  
    & 6002  $\pm$ 430   &  5059 $\pm$ 499   & 1766 $\pm$ 803    &  1956 $\pm$ 949     & 1567 $\pm$ 887      & 1384 $\pm$ 697   & 2955.7  \\
    & SAC-TIRL-VAED 
    & 5957  $\pm$ 804   &  5034 $\pm$ 525   & 2054 $\pm$ 784    &  2207 $\pm$ 752     & 1459 $\pm$ 467      & 1646 $\pm$ 851   & 3059.5  \\
    \noalign{\hrule height 0.8pt}
    
    \multirow{5}{*}{\specialcell{\textbf{Hopper}\\(state-dim: 11)\\$\epsilon$: 0.075}}  
    & SAC 
    & \textbf{3476 $\pm$ 163}    &  \textbf{2967 $\pm$ 414}   & 1570 $\pm$ 261    & 1557 $\pm$ 325    &  1098 $\pm$ 194       & 1028 $\pm$ 279   & 1949.3 \\
    & SAC-TIRL-BDR 
    & \textbf{3516 $\pm$ 90}     &  \textbf{3015 $\pm$ 427}   & 2620 $\pm$ 695    & 2320  $\pm$  760  &  1687 $\pm$ 432       & 1774 $\pm$ 935   & 2488.7 \\
    & \textbf{SAC-TIRL-VQ} 
    & 3253 $\pm$ 182    &  \textbf{3123 $\pm$ 288}   & \textbf{3177 $\pm$ 218}    & \textbf{3188 $\pm$ 300}    & \textbf{2500 $\pm$ 46}         & \textbf{2997 $\pm$ 595}   & \textbf{3039.7} \\
    & SAC-TIRL-AED 
    & 3195 $\pm$ 409    & 2303$\pm$383      & 1729 $\pm$ 325    &  1956 $\pm$ 221   & 1461 $\pm$ 47         & 1499 $\pm$ 565   & 2023.8 \\
    & SAC-TIRL-VAED 
    & 3189  $\pm$ 665   & 2914 $\pm$ 289    & 1749 $\pm$ 176    &  1606 $\pm$ 277   & 1593 $\pm$ 367        & 1704 $\pm$ 291   & 2125.8 \\
    \noalign{\hrule height 0.8pt}

    \multirow{5}{*}{\specialcell{\textbf{Walker2d}\\(state-dim: 17)\\$\epsilon$: 0.05}}  
    & SAC      
    & \textbf{6242 $\pm$ 383}    &  \textbf{6067 $\pm$ 433}   & 3700 $\pm$ 655    &  3524 $\pm$ 600   &   3854 $\pm$ 1359     & 1782 $\pm$ 818   & 4194.8  \\
    & \textbf{SAC-TIRL-BDR} 
    & \textbf{6266 $\pm$ 417}    &  \textbf{6243 $\pm$ 392}   & \textbf{6142 $\pm$ 387}    &  \textbf{6144 $\pm$ 401}   &   \textbf{5651 $\pm$ 688}      & \textbf{5968 $\pm$ 156}   & \textbf{6069.0}  \\
    & SAC-TIRL-VQ 
    & 5455 $\pm$ 459    &  5298 $\pm$ 479   & 5400 $\pm$ 547    &   5333 $\pm$ 277  &   5225 $\pm$ 607      & 5211 $\pm$ 453   & 5320.3 \\
    & SAC-TIRL-AED 
    & \textbf{6211  $\pm$  367}  &  \textbf{5938  $\pm$ 458}  & 4374 $\pm$ 797    &  4215 $\pm$ 1291  &   4909 $\pm$ 1122     & 3280 $\pm$ 1332  & 4821.2 \\
    & SAC-TIRL-VAED 
    & \textbf{6184  $\pm$ 210}   &  5704  $\pm$ 189  & 4596 $\pm$ 635    &  3975 $\pm$ 1090  &   4690 $\pm$ 1260     & 3240 $\pm$ 1589  & 4731.5 \\
    \noalign{\hrule height 0.8pt}
    
    \multirow{5}{*}{\specialcell{\textbf{Pendulum}\\(state-dim: 4)\\$\epsilon$: 0.3}}    
    & SAC      
    & \textbf{1000 $\pm$ 0}      &  916 $\pm$ 169    & 638 $\pm$ 304     &  736 $\pm$ 317    &  89 $\pm$ 76          & 188 $\pm$ 94    & 594.5 \\
    & \textbf{SAC-TIRL-BDR} 
    & \textbf{1000 $\pm$ 0}      &  \textbf{1000 $\pm$ 0}     & \textbf{953 $\pm$ 66}      &  \textbf{963 $\pm$ 52}     &  434 $\pm$ 94         & \textbf{789 $\pm$ 97}    & \textbf{856.5} \\
    & SAC-TIRL-VQ 
    & \textbf{1000 $\pm$ 0}      &  \textbf{1000 $\pm$ 0}     & 738 $\pm$ 230     &  814 $\pm$ 263    &  \textbf{797 $\pm$ 134}        & 703 $\pm$ 95    & \textbf{842.0} \\
    & SAC-TIRL-AED 
    & \textbf{1000 $\pm$ 0}      &  \textbf{1000 $\pm$ 0}     & \textbf{949 $\pm$ 40}      &  869 $\pm$ 155    &  474 $\pm$ 68         & 489 $\pm$ 65    & 796.8 \\
    & SAC-TIRL-VAED 
    & \textbf{1000 $\pm$ 0}      &  \textbf{1000 $\pm$ 0}     & 775 $\pm$ 285     &  756 $\pm$ 388    &  499 $\pm$ 92         & 534 $\pm$ 99    & 760.7 \\
    \noalign{\hrule height 0.8pt}
    
    \multirow{5}{*}{\specialcell{\textbf{Reacher}\\(state-dim: 11)\\$\epsilon$: 1.0}}
    & SAC   
    & \textbf{-3.68 $\pm$ 0.07}  &  -7.77 $\pm$ 0.66 & -32.31 $\pm$ 2.0      &  -29.59 $\pm$ 2.51    & -21.52  $\pm$ 0.53 &  -21.36 $\pm$ 0.72   & -19.4 \\
    & SAC-TIRL-BDR 
    & \textbf{-3.76 $\pm$ 0.03}  &  \textbf{-6.99 $\pm$ 0.32} & -26.95 $\pm$ 1.76     &  -24.93 $\pm$ 1.35    &  -17.58 $\pm$ 2.62 &  -17.30 $\pm$ 0.52   & -16.3  \\
    & SAC-TIRL-VQ 
    & -4.46 $\pm$ 0.03  &  -7.57 $\pm$ 0.17 & -19.02 $\pm$ 1.87     &  -18.60 $\pm$ 1.29    &  -13.0  $\pm$ 0.39 &  \textbf{-14.1  $\pm$ 1.21}   & -12.8  \\
    & SAC-TIRL-AED 
    & \textbf{-3.67  $\pm$ 0.09} &  -8.22 $\pm$ 0.61 & -21.8 $\pm$ 3.0       &  -21.46 $\pm$ 2.13    &  -11.49 $\pm$ 1.73 &  -14.90 $\pm$ 0.98   & -13.6  \\
    & \textbf{SAC-TIRL-VAED} 
    & \textbf{-3.68  $\pm$ 0.10} &  \textbf{-7.04 $\pm$ 0.34} & \textbf{-16.59 $\pm$ 1.61}     &  \textbf{-16.51 $\pm$ 1.59}    &  \textbf{-10.71 $\pm$ 1.56} &  \textbf{-14.72 $\pm$ 1.41}   & \textbf{-11.5} \\
    \noalign{\hrule height 0.8pt}
    \end{tabular}
    \vspace{-0.15in}
\end{table*}
\noindent
\textbf{Input Transformations for Robust RL} 
\begin{itemize}[leftmargin=*]
    \item TIRL-BDR: We apply the BDR transformation to the input of the policy network of SAC, while leaving the input of the $Q$ networks unchanged. Note that BDR is applied in \textit{both} training (in a regular environment) and testing (with adversaries). The hyperparameter $bW$ is searched in $\{0.1, 0.25, 0.4, 0.5\}$ for Reacher, and $\{0.05, 0.1, 0.15, 0.2\}$ for the other environments.
    
    \item TIRL-VQ: Similar to BDR, the VQ transformation is applied only to the input of the policy network and is used during both training and testing. The codebook is learned concurrently with the agent while training by optimizing Eq.(~\ref{eq:vq_learning}); during testing, the learned codebook is fixed. The codebook size $K$ is searched in $\{2\times 10^5, 5\times 10^5, 10^6\}$ for Ant, Hopper, and Walker2d, and $\{5\times 10^4, 1\times 10^5, 2\times 10^5, 3\times 10^5, 5\times 10^5\}$ for the other environments.
    
    \item TIRL-AED: We use an autoencoder-based denoiser to reconstruct an original state from a perturbed state. To train the denoiser, we collect a set of $N$ states $\{s_i\}_{i=1}^N$ from a pretrained policy's replay buffer, then use \textit{Min Q} attacks to generate perturbed states  $\{\tilde{s}_i\}_{i=1}^N$. The autoencoder is trained to reconstruct the original states by optimizing the mean square error: $\|s - D_{\omega_2}(E_{\omega_1}(\tilde{s})\|_2^2$, where $E_{\omega_1}$ and $D_{\omega_2}$ are the encoder and decoder, respectively. Note that the denoiser is only used to denoise the perturbed inputs of the policy network during testing. We use the Adam optimizer \cite{kingma2014method} for training the denoiser, with a learning rate of $3\times 10^{-4}$. For the \textit{Min Q} attacker, we search the $\epsilon$-radius in $\{0.25, 0.3, 0.35, 0.4\}$ for Inverted Pendulum, and $\{0.05, 0.1, 0.15, 0.2, 0.25\}$ for the other environments.
    
    \item TIRL-VAED: We follow a similar setting as TIRL-AED, except for the training objective. We utilize the variational autoencoder \cite{kingma2013auto} objective: $-\beta D_{KL}(E_{\omega_1}(z|\tilde{s}) \| p(z))+\mathbb{E}_{E_{\omega_1}(z|\tilde{s})}\left[\log D_{\omega_2}(s|z) \right]$, where $p(z)$ is the standard normal distribution, and $\beta$ is set to $1\times 10^{-4}$. We also search the $\epsilon$-radius in the same range as TIRL-AED.
\end{itemize}
We summarize the best hyperparameters for transformations across environments in the Table \ref{tab:main_hyperparam}.

\begin{table*}[h]
\renewcommand{\arraystretch}{1.3}
\centering
\caption{Average episode return $\pm$ standard deviation of SAC agents and SAC-TIRL with various transformation defenses in the white-box setting. Results are averaged over 5 runs, with evaluations conducted over 50 episodes per run. Following \cite{kostrikov2021offline}, we \textbf{bold} all scores within 5 percent of the maximum in each attack and environment ($\geq$ 0.95 $\cdot$ max).}
\label{tab:white_box}
\begin{tabular}{l@{\hspace{4pt}}|l@{\hspace{4pt}}|c@{\hspace{6pt}}|c@{\hspace{6pt}}c@{\hspace{6pt}}c@{\hspace{6pt}}c@{\hspace{6pt}}c@{\hspace{6pt}}c@{\hspace{6pt}}}
    \noalign{\hrule height 0.8pt}
    \textbf{Environment}   & \textbf{Method}  &   \textbf{Natural Reward} & \textbf{Random} & \textbf{Action Diff}   & \textbf{Min Q}   &  \textbf{RS} & \textbf{PA-AD} & \textbf{Average}\\
    \noalign{\hrule height 0.8pt}
    \multirow{5}{*}{\specialcell{\textbf{Ant}\\(state-dim: 111)\\$\epsilon$: 0.15}}   
    & SAC   
    & \textbf{6859  $\pm$ 126}   &  \textbf{5637 $\pm$ 293}   & 454  $\pm$ 335    &  370 $\pm$ 279      & 463 $\pm$ 216       &  -962 $\pm$ 341  & 2136.8  \\   
    & SAC-TIRL-BDR  
    & \textbf{6543  $\pm$ 506}   &  \textbf{5445 $\pm$ 477}   & 877 $\pm$ 54      &  767 $\pm$ 174      & 1073 $\pm$ 375      &  719  $\pm$ 513  & 2570.7  \\
    & \textbf{SAC-TIRL-VQ}   
    & 4785  $\pm$ 164   &  4737 $\pm$ 205   & \textbf{4699 $\pm$ 308}    &  \textbf{4456 $\pm$ 236}     & \textbf{4379 $\pm$ 129}      & \textbf{4103$ \pm$ 161}   & \textbf{4526.5}  \\
    & SAC-TIRL-AED  
    & 6002  $\pm$ 430   &  5059 $\pm$ 499   & 721 $\pm$ 153     &  819  $\pm$ 153     &  479 $\pm$ 393      &  -154 $\pm$ 302  & 2154.3  \\
    & SAC-TIRL-VAED 
    & 5957  $\pm$ 804   &  5034 $\pm$ 525   & 799 $\pm$ 395     &  864  $\pm$  341    &  590 $\pm$ 314      & -150  $\pm$ 308  & 2182.3  \\
    \noalign{\hrule height 0.8pt}
    
    \multirow{5}{*}{\specialcell{\textbf{Hopper}\\(state-dim: 11)\\$\epsilon$: 0.075}}  
    & SAC 
    & \textbf{3476 $\pm$ 163}    &  \textbf{2967 $\pm$ 414}   & 1570 $\pm$ 261    & 1557 $\pm$ 325    &  1098 $\pm$ 194       & 1028 $\pm$ 279   & 1949.3 \\
    & SAC-TIRL-BDR 
    & \textbf{3516 $\pm$ 90}     &  \textbf{3015 $\pm$ 427}   & 2620 $\pm$ 695    & 2300  $\pm$ 634   &  1613 $\pm$ 432        & 1521 $\pm$ 811   & 2430.8 \\
    & \textbf{SAC-TIRL-VQ} 
    & 3253 $\pm$ 182    &  \textbf{3123 $\pm$ 288}   & \textbf{2947 $\pm$ 334}    & \textbf{2706 $\pm$ 129}    & \textbf{2369 $\pm$ 301}        & \textbf{2634 $\pm$ 491}   & \textbf{2838.7} \\
    & SAC-TIRL-AED 
    & 3195 $\pm$ 409    & 2303$\pm$383      &  1374 $\pm$ 392   &  1527 $\pm$ 307   & 1234 $\pm$ 471         & 1176 $\pm$ 182   & 1801.5 \\
    & SAC-TIRL-VAED 
    & 3189  $\pm$ 665   & 2914 $\pm$ 289    & 1445 $\pm$ 396    & 1864  $\pm$ 299   & 1446 $\pm$ 213        & 1097 $\pm$ 241   & 1992.5 \\
    \noalign{\hrule height 0.8pt}

    \multirow{5}{*}{\specialcell{\textbf{Walker2d}\\(state-dim: 17)\\$\epsilon$: 0.05}}  
    & SAC      
    & \textbf{6242 $\pm$ 383}    &  \textbf{6067 $\pm$ 433}   & 3700 $\pm$ 655    &  3524 $\pm$ 600   &   3854 $\pm$ 1359     & 1782 $\pm$ 818   & 4194.8  \\
    & \textbf{SAC-TIRL-BDR} 
    & \textbf{6266 $\pm$ 417}    &  \textbf{6243 $\pm$ 392}   &  5180 $\pm$ 908    &  \textbf{5265 $\pm$ 1028}   &   \textbf{5198 $\pm$ 550}      & 4019 $\pm$ 1558   & \textbf{5361.8}  \\
    & SAC-TIRL-VQ 
    & 5455 $\pm$ 459    &  5298 $\pm$ 479   & \textbf{5560 $\pm$ 282}    &  4823  $\pm$ 524  &   \textbf{5370 $\pm$ 349}      & \textbf{5193 $\pm$ 558}   & 5283.2 \\
    & SAC-TIRL-AED 
    & \textbf{6211  $\pm$  367}  &  \textbf{5938  $\pm$ 458}  & 4117 $\pm$  1130   & 3942  $\pm$ 927  &   4876 $\pm$ 947     & 3011 $\pm$ 633  & 4682.5 \\
    & SAC-TIRL-VAED 
    & \textbf{6184  $\pm$ 210}   &  5704  $\pm$ 189  & 4158 $\pm$ 1291    & 3752  $\pm$ 1484  &   4639 $\pm$ 1260     & 3143 $\pm$ 858  & 4596.7 \\
    \noalign{\hrule height 0.8pt}
    
    \multirow{5}{*}{\specialcell{\textbf{Pendulum}\\(state-dim: 4)\\$\epsilon$: 0.3}}    
    & SAC      
    & \textbf{1000 $\pm$ 0}      &  916 $\pm$ 169    & 638 $\pm$ 304     &  736 $\pm$ 317    &  89 $\pm$ 76          & 188 $\pm$ 94    & 594.5 \\
    & SAC-TIRL-BDR 
    & \textbf{1000 $\pm$ 0}      &  \textbf{1000 $\pm$ 0}     &  \textbf{934 $\pm$  47}    & \textbf{946  $\pm$  77}    &  272 $\pm$ 235         & 223 $\pm$  97   & 729.2 \\
    & \textbf{SAC-TIRL-VQ} 
    & \textbf{1000 $\pm$ 0}      &  \textbf{1000 $\pm$ 0}     &  820 $\pm$  181   & 744  $\pm$ 338  &  \textbf{621 $\pm$ 360}        & \textbf{597 $\pm$ 97}    & \textbf{797.0} \\
    & SAC-TIRL-AED 
    & \textbf{1000 $\pm$ 0}      &  \textbf{1000 $\pm$ 0}     & 386 $\pm$ 436      &  398 $\pm$ 428    & 324  $\pm$ 125         & 248 $\pm$ 111    & 559.3 \\
    & SAC-TIRL-VAED 
    & \textbf{1000 $\pm$ 0}      &  \textbf{1000 $\pm$ 0}     & 245 $\pm$ 29     &  241 $\pm$ 27    &  240 $\pm$ 15         & 121 $\pm$  15   & 474.5 \\
    \noalign{\hrule height 0.8pt}
    
    \multirow{5}{*}{\specialcell{\textbf{Reacher}\\(state-dim: 11)\\$\epsilon$: 1.0}} 
    & SAC   
    & \textbf{-3.68 $\pm$ 0.07}  &  -7.77 $\pm$ 0.66 & -32.31 $\pm$ 2.0      &  -29.59 $\pm$ 2.51    & -21.52  $\pm$ 0.53 &  -21.36 $\pm$ 0.72   & -19.4 \\
    & SAC-TIRL-BDR 
    & \textbf{-3.76 $\pm$ 0.03}  &  \textbf{-6.99 $\pm$ 0.32} & -27.37 $\pm$  1.97   &   -25.37 $\pm$ 0.87    &  -18.12 $\pm$ 1.67   & -19.91 $\pm$ 2.03  &   -16.9 \\
    & \textbf{SAC-TIRL-VQ} 
    & -4.46 $\pm$ 0.03  &  -7.57 $\pm$ 0.17 & \textbf{-15.83 $\pm$ 0.05}     &  \textbf{-16.25 $\pm$ 1.25}   &    -15.2 $\pm$ 1.29    &  \textbf{-14.90  $\pm$ 2.16} & \textbf{-12.4}  \\
    & SAC-TIRL-AED 
    & \textbf{-3.67  $\pm$ 0.09} &  -8.22 $\pm$ 0.61 & -22.15 $\pm$ 1.74      &  -23.08 $\pm$ 1.88    &   \textbf{-13.21 $\pm$ 2.06} &  -15.90 $\pm$ 1.22   & -14.4  \\
    & SAC-TIRL-VAED 
    & \textbf{-3.68  $\pm$ 0.10} &  \textbf{-7.04 $\pm$ 0.34} &  -18.2 $\pm$ 1.24     &  -19.5 $\pm$ 2.25    &  -16.01 $\pm$ 1.22   &  \textbf{-14.21 $\pm$ 1.64}   & -13.1 \\
    \noalign{\hrule height 0.8pt}
    \end{tabular}
    \vspace{-0.15in}
\end{table*}

\subsection{Evaluation with Gray-box Attacks}\label{subsec:graybox}
In this scenario, adversaries can access the policy architecture and its parameters but not the applied transformations. When generating perturbations, the clean state is fed directly into the policy network, bypassing transformations. For example, in the \textit{Action Diff} attack, the adversarial state $\tilde{s}$ is generated using the formula: $\tilde{s}^{k+1} = \tilde{s}^{k} + \eta \; \text{proj} [\nabla_{\tilde{s}^k} D_{KL}(\pi_{\theta}(\cdot|s)||\pi_{\theta}(\cdot|\tilde{s}^k)]$, ignoring any transformations applied during policy training. \textit{Min Q} and \textit{RS} are analogous to \textit{Action Diff}. In RL-based attacks like PA-AD, the transformation is also excluded from the pre-trained policy when learning the adversarial policy.

The performance of both the vanilla SAC agent and the TIRL with various transformations under different attacks is presented in Table \ref{tab:gray_box}. The results illustrate the effectiveness of the investigated transformations in mitigating the impact of attacks. The TIRL-BDR improves performance by $31\%$ over the vanilla SAC across five environments, achieving the highest robustness in the Walker2d and Inverted Pendulum environments at $45\%$ and $44\%$, respectively. The TIRL-VQ significantly enhances robustness by $55\%$ over the vanilla SAC, with the most substantial improvements observed in the Ant ($117\%$) and Hopper ($56\%$) environment among transformations. In the Reacher environment, the TIRL-VAED excels with the highest performance, reaching $40\%$. When comparing the TIRL-AED and TIRL-VAED, both demonstrate comparable performance. Overall, in the gray-box setting, the TIRL-VQ transformation appears to be the most effective among the transformations considered. However, it is essential to note that applying VQ impacts natural returns due to the loss of information in quantized states. This highlights the trade-off between balancing robustness and maintaining natural performance.

\newcommand{\DECREASE}[1]{\textcolor{blue}{#1}}

\begin{table*}[ht]
\renewcommand{\arraystretch}{1.3}
\centering
\setlength{\tabcolsep}{5.5pt}
\caption{Average episode return $\pm$ standard deviation of different defenses. Results are averaged over 5 runs, with evaluations conducted over 50 episodes per run. Following \cite{kostrikov2021offline}, we \textbf{bold} all scores within 5 percent of the maximum in each attack and environment ($\geq$ 0.95 $\cdot$ max). Note that \textit{Min Q} is not applicable for PA-ATLA since it does not contain a $Q$ network.}
\label{tab:comparison}
\begin{tabular}{l@{\hspace{4pt}}|l@{\hspace{4pt}}|c@{\hspace{6pt}}|c@{\hspace{6pt}}c@{\hspace{6pt}}c@{\hspace{6pt}}c@{\hspace{6pt}}c@{\hspace{6pt}}c@{\hspace{6pt}}}
    \noalign{\hrule height 0.8pt}
    \textbf{Environment}   & \textbf{Method}  &   \textbf{Natural Reward} & \textbf{Random} & \textbf{Action Diff}   & \textbf{Min Q}   &  \textbf{RS} & \textbf{PA-AD} & \textbf{Average}\\
    \noalign{\hrule height 0.8pt}
    \multirow{4}{*}{\specialcell{\textbf{Ant}\\(state-dim: 111)\\$\epsilon$: 0.15}}    
    & SAC-TIRL-BDR  
    & \textbf{6543  $\pm$ 506}   &  \textbf{5445 $\pm$ 477}   & 877 $\pm$ 54      &  767 $\pm$ 174      & 1073 $\pm$ 375      &  719  $\pm$ 513  & 2570.7  \\
    & \textbf{SAC-TIRL-VQ}   
    & 4785  $\pm$ 164   &  4737 $\pm$ 205   & 4699 $\pm$ 308    &  \textbf{4456 $\pm$ 236}     & \textbf{4379 $\pm$ 129}      & \textbf{4103$ \pm$ 161}   & \textbf{4526.5}  \\
    & SAC-SA  
    & \textbf{6626  $\pm$ 675}   &  \textbf{5263 $\pm$ 1657}  & 4357 $\pm$ 1734   &  3735  $\pm$ 1647   &  3414 $\pm$ 850      &  1021 $\pm$ 943  & 4069.3  \\
    & PA-ATLA  
    & 5469  $\pm$ 106   &  \textbf{5469 $\pm$ 158}  & \textbf{5328 $\pm$ 169}   &  -   &  4124 $\pm$ 291      &  2986 $\pm$ 864  &  \textbf{4675.2} \\
    \noalign{\hrule height 0.8pt}
    
    \multirow{4}{*}{\specialcell{\textbf{Hopper}\\(state-dim: 11)\\$\epsilon$: 0.075}}  
    & SAC-TIRL-BDR 
    & 3516 $\pm$ 90     &  3015 $\pm$ 427   & 2620 $\pm$ 695    & 2300  $\pm$ 634   &  1613 $\pm$ 432        & 1521 $\pm$ 811   & 2430.8 \\
    & SAC-TIRL-VQ 
    & 3253 $\pm$ 182    &  3123 $\pm$ 288   & 2947 $\pm$ 334    & 2706 $\pm$ 129    & 2369 $\pm$ 301        & \textbf{2634 $\pm$ 491}   & 2838.7 \\
    & \textbf{SAC-SA} 
    & \textbf{3652 $\pm$ 359}    & \textbf{3522 $\pm$ 375}    & \textbf{3394 $\pm$ 675}    &  \textbf{2987 $\pm$ 359}   &  \textbf{2869 $\pm$ 862}   & \textbf{2504 $\pm$ 653}         & \textbf{3154.7} \\
    & PA-ATLA  
    & 3449  $\pm$ 237   &  3325 $\pm$ 239  & 3145 $\pm$ 546   &  -   &  \textbf{3002 $\pm$ 129}      &  \textbf{2521 $\pm$ 325}  &  \textbf{3086.6} \\
    \noalign{\hrule height 0.8pt}

    \multirow{4}{*}{\specialcell{\textbf{Walker2d}\\(state-dim: 17)\\$\epsilon$: 0.05}}  
    & SAC-TIRL-BDR 
    & \textbf{6266 $\pm$ 417}    &  \textbf{6243 $\pm$ 392}   &  5180 $\pm$ 908    &  \textbf{5265 $\pm$ 1028}   &   \textbf{5198 $\pm$ 550}      & 4019 $\pm$ 1558   & \textbf{5361.8}  \\
    & SAC-TIRL-VQ 
    & 5455 $\pm$ 459    &  5298 $\pm$ 479   & \textbf{5560 $\pm$ 282}    &  4823  $\pm$ 524  &   \textbf{5370 $\pm$ 349}      & \textbf{5193 $\pm$ 558}   & \textbf{5283.2} \\
    & \textbf{SAC-SA} 
    & 5895  $\pm$  305  &  5884  $\pm$ 419  & \textbf{5502 $\pm$  499}   & \textbf{5275 $\pm$  234}   & \textbf{5124  $\pm$ 383}  &   4505 $\pm$ 684  & \textbf{5364.3} \\
    & PA-ATLA  
    & 4178  $\pm$ 529   &  4129 $\pm$ 78  & 4024 $\pm$ 572   &  -   &  3966 $\pm$ 307      &  2248 $\pm$ 131  &  3709.0 \\
    \noalign{\hrule height 0.8pt}
    \end{tabular}
    \vspace{-0.15in}
\end{table*}

\begin{figure*}[t]
    \begin{subfigure}{\textwidth}
        \centering
        \includegraphics[width=\textwidth]{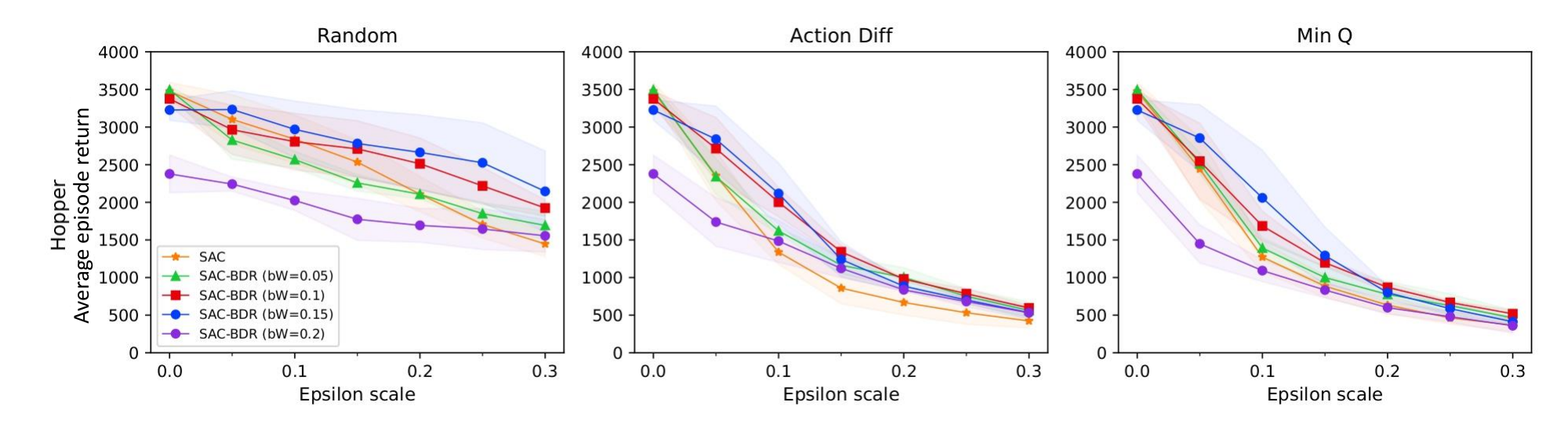}
    \end{subfigure}
    \caption{\textbf{The ablation study of different bin width ($bW$)} for BDR transformation in Hopper environment. We evaluate the robustness under different attacks with various $\epsilon$ scales.}
    \label{fig:bdr_ablation}
    \vspace{-0.16in}
\end{figure*}
\begin{figure*}[t]
    \begin{subfigure}{\textwidth}
        \centering
        \includegraphics[width=\textwidth]{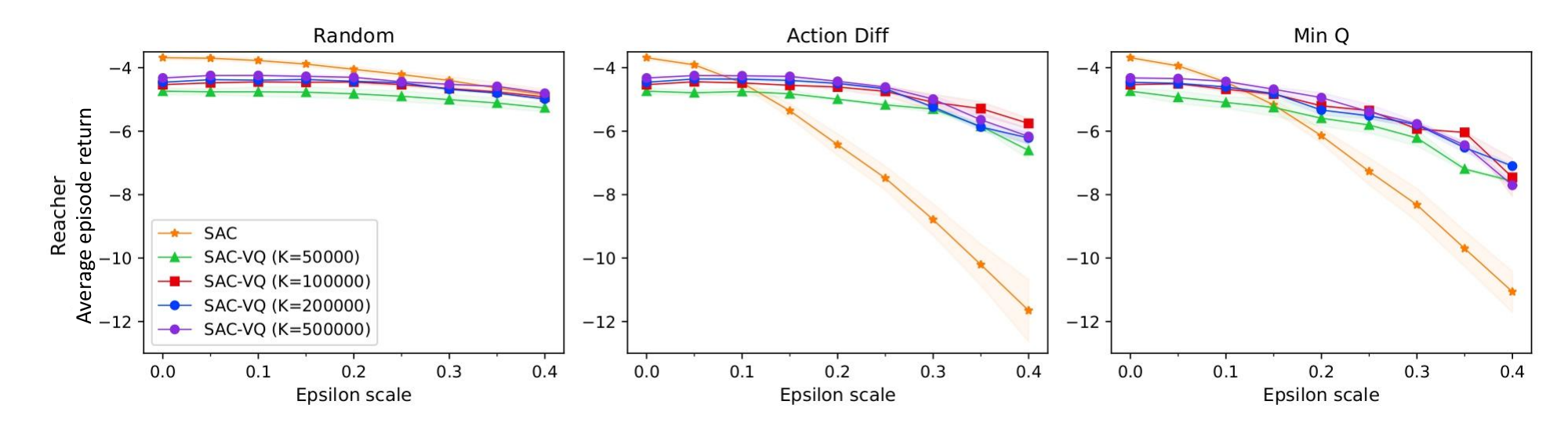}
    \end{subfigure}

    \caption{\textbf{The abluation study of different codebook size ($K$)} for VQ transformation in Reacher environment. The agents are evaluated under different attacks with various $\epsilon$ scales.}
    \label{fig:vanilla_vq_ablation}
    \vspace{-0.16in}
\end{figure*}

\subsection{Evaluation with White-box Attacks}\label{subsec:whitebox}
While the previous experiment demonstrates that input transformation defenses are relatively effective in mitigating adversarial attacks, we now conduct experiments using white-box attacks in a more challenging setting. In this scenario, adversaries can access the policy architecture, parameters, and applied transformations. Specifically, in the \textit{Action Diff}, \textit{Min Q}, and \textit{RS} attacks, adversaries are allowed to backpropagate through the transformations while generating perturbations. For example, the \textit{Action Diff} generates an adversarial state as follows: $\tilde{s}^{k+1} = \tilde{s}^{k} + \eta \; \text{proj} [\nabla_{\tilde{s}^k} D_{KL}(\pi_{\theta}(\cdot|\mathcal{T}(s))||\pi_{\theta}(\cdot|\mathcal{T}(\tilde{s}^k))]$. Similarly, RL-based adversaries can use the complete pretrained policy, including its transformations, in RL-based attacks to train the adversarial policy.

Table~\ref{tab:white_box} presents the performance results in this scenario. The results show that TIRL-VQ maintains the highest robustness among all transformations, achieving a $50\%$ improvement over vanilla SAC.  Although TIRL-BDR's improvement decreases, it still partially mitigates the attacks' effects, showing a $20\%$ improvement across five environments. In contrast, the autoencoder-style denoising approach exhibits a performance decline in this setting, resulting in outcomes similar to vanilla SAC. This decline is because the denoisers are composed of deep neural network components, which adversaries can easily exploit to generate more robust perturbations. Additionally, we observe that TIRL-VQ performs best in 4 out of 5 environments in the white-box setting, compared to 2 out of 5 environments in the gray-box setting. This indicates that VQ's performance hardly deteriorates as the adversary's strength increases. 

\subsection{Comparison with prior work}

In this experiment, we compare our defenses with prior robust training-based methods, including SAC-SA \cite{zhang2020robust} and PA-ATLA \cite{sun2021strongest}. We conduct experiments on the Ant, Hopper, and Walker2d environments, which were also used in previous works. The results are presented in Table \ref{tab:comparison}. In the Ant environment, we observe that SAC-TIRL-VQ performs comparably to PA-ATLA and better than SAC-SA. In the Walker2d environment, SAC-TIRL-VQ and TIRL-BDR are comparable to SAC-SA and outperform PA-ATLA. Across all three environments, SAC-TIRL-VQ achieves an average return of $4216.1$, SAC-SA achieves an average return of $4196.1$, and PA-ATLA achieves an average return of $3823.6$. Overall, SAC-TIRL-VQ shows comparable improvement to SAC-SA across environments, with an approximate 52\% improvement compared to vanilla SAC. It is worth noting that, unlike SAC-SA and PA-ATLA, we do not use adversarial examples during training. Moreover, our method only modifies the model's input, making it possible to combine with robust training-based defenses for further robustness enhancement. We leave this investigation for future work.

\subsection{Ablation experiments}
In this section, we conduct an ablation study on hyperparameters for TIRL-BDR and TIRL-VQ transformations. We first train the agents with different hyperparameters. Subsequently, with each pretrained model, we evaluate robustness as a function of adversary strength (\ie, the $\epsilon$-radius) for each of the three attackers: \textit{Random}, \textit{Action Diff}, and \textit{Min Q}. 

Fig.~\ref{fig:bdr_ablation} illustrates the robustness of TIRL-BDR in the Hopper environment with varying values of bin width ($bW$). As observed, if $bW$ is small (\eg, 0.05), it resembles the robustness of vanilla SAC, particularly noticeable in \textit{Action Diff} and \textit{Min Q} attacks. This is because a smaller $bW$ essentially reflects the original state space, thus having less impact on countering perturbations caused by attacks. However, a value that is too large, $bW$, dramatically drops natural performance due to inaccurate representations.

Fig.~\ref{fig:vanilla_vq_ablation} showcases the results for various codebook sizes of the VQ transformation in the Reacher environment. Across all considered values, robustness in policy performance remains consistently strong. However, our primary objective is to identify smaller codebook sizes that yield higher natural rewards. This choice is motivated by the increasing computational cost of larger codebook sizes in the VQ transformation. Notably, in our experiments on the Hopper environment, the wall-clock time for training SAC-VQ with $K=5\times 10^5$ is approximately 2.5 times longer compared to SAC-BDR. This cost differential becomes even more pronounced in higher-dimensional environments. Consequently, when selecting hyperparameters for VQ, we recommend commencing with smaller values and incrementally adjusting them until achieving a satisfactory balance between robustness and computational efficiency. 

\subsection{Computational Cost Comparison}
\begin{table}[ht]
\renewcommand{\arraystretch}{1.3}
\centering
\caption{Average training and testing times in the Ant environment. Note that only Bounded Transformation is used during RL agent training, while the Autoencoder-styled Denoising is applied after the agent's training is complete.}
\begin{tabular}{|l|cc|}
\hline
\multicolumn{1}{|c|}{\multirow{3}{*}{\textbf{Method}}} & \multicolumn{2}{c|}{\textbf{Running time}}                                                                                                                            \\ \cline{2-3} 
\multicolumn{1}{|c|}{}                        & \multicolumn{1}{c|}{\begin{tabular}[c]{@{}c@{}}\textbf{Training}\\ (hours per $10^6$ steps)\end{tabular}} & \begin{tabular}[c]{@{}c@{}}\textbf{Testing}\\ (seconds per 50 episodes)\end{tabular} \\ \hline
SAC                                           & \multicolumn{1}{c|}{5.28}                                                                & 91                                                              \\ \hline
SAC-TIRL-BDR                                  & \multicolumn{1}{c|}{5.38}                                                                & 91.5                                                              \\ \hline
SAC-TIRL-VQ                                   & \multicolumn{1}{c|}{10.54}                                                               & 109.5                                                              \\ \hline
SAC-TIRL-AED                                  & \multicolumn{1}{c|}{-}                                                                   & 92.5                                                              \\ \hline
SAC-TIRL-VAED                                 & \multicolumn{1}{c|}{-}                                                                   & 92.5                                                              \\ \hline
\end{tabular}
\label{tab:computation_cost}
\vspace{-0.2in}
\end{table}

We compare the training and testing times for different transformations on a single machine equipped with a Tesla V100 16GB GPU. Note that TIRL-AED and TIRL-VAED are trained after completing the training of RL agents, which takes approximately 15-20 minutes for training the denoiser. The results are presented in Table \ref{tab:computation_cost}. Although TIRL-VQ is effective in mitigating adversarial attacks, its training time with SAC is longer, primarily due to the K-means algorithm. This poses a limitation when extending our method to high-dimensional states, such as images. Future investigations could explore more advanced variants of VQ to optimize efficiency further and expand the scope of potential applications.

\section{Conclusions}
Reinforcement Learning (RL) agents demonstrating proficiency in a training environment exhibit vulnerability to adversarial perturbations in input observations during deployment. This underscores the importance of building a robust agent before its real-world deployment. To alleviate the challenging point, prior works focus on developing robust training-based procedures, encompassing efforts to fortify the deep neural network component's robustness or subject the agent to adversarial training against potent attacks. In this work, we proposed a novel method referred to as \textit{Transformed Input-robust RL (TIRL)}, which explores another avenue to mitigate the impact of adversaries by employing input transformation-based defenses. Specifically, we introduced two principles for applying transformation-based defenses in learning robust RL agents: \textit{(1) autoencoder-styled denoising} to reconstruct the original state and \textit{(2) bounded transformations (bit-depth reduction and vector quantization (VQ))} to achieve close transformed inputs. The transformations are applied to the state before feeding them into the policy network. Extensive experiments on multiple \mujoco environments demonstrated that input transformation-based defenses, \ie, VQ, defend against several adversaries in the state observations.

In this paper, we focus primarily on low-dimensional environments due to the high computational cost of applying VQ to high-dimensional environments like those in Atari games. Future research can address this challenge by developing more advanced and efficient VQ techniques. Additionally, our proposed TIRL method holds the potential to synergistically complement other robust training-based defenses, significantly enhancing the overall robustness of RL agents. This direction opens exciting opportunities for developing more resilient RL systems capable of operating effectively in complex and adversarial environments.

\bibliographystyle{ieeetr}
\bibliography{references}

\begin{thebibliography}{10}

\bibitem{mnih2015human}
V.~Mnih, K.~Kavukcuoglu, D.~Silver, A.~A. Rusu, J.~Veness, M.~G. Bellemare,
  A.~Graves, M.~Riedmiller, A.~K. Fidjeland, G.~Ostrovski, {\em et~al.},
  ``Human-level control through deep reinforcement learning,'' {\em Nature},
  pp.~529--533, 2015.

\bibitem{schulman2015trust}
J.~Schulman, S.~Levine, P.~Abbeel, M.~Jordan, and P.~Moritz, ``Trust region
  policy optimization,'' in {\em ICML}, pp.~1889--1897, 2015.

\bibitem{silver2016mastering}
D.~Silver, A.~Huang, C.~J. Maddison, A.~Guez, L.~Sifre, G.~Van Den~Driessche,
  J.~Schrittwieser, I.~Antonoglou, V.~Panneershelvam, M.~Lanctot, {\em et~al.},
  ``Mastering the game of go with deep neural networks and tree search,'' {\em
  Nature}, pp.~484--489, 2016.

\bibitem{fujimoto2018addressing}
S.~Fujimoto, H.~Hoof, and D.~Meger, ``Addressing function approximation error
  in actor-critic methods,'' in {\em ICML}, pp.~1587--1596, 2018.

\bibitem{haarnoja2018soft}
T.~Haarnoja, A.~Zhou, P.~Abbeel, and S.~Levine, ``Soft actor-critic: Off-policy
  maximum entropy deep reinforcement learning with a stochastic actor,'' in
  {\em ICML}, pp.~1861--1870, 2018.

\bibitem{luu2021hindsight}
T.~M. Luu and C.~D. Yoo, ``Hindsight goal ranking on replay buffer for sparse
  reward environment,'' {\em IEEE Access}, vol.~9, pp.~51996--52007, 2021.

\bibitem{kostrikov2021offline}
I.~Kostrikov, A.~Nair, and S.~Levine, ``Offline reinforcement learning with
  implicit q-learning,'' {\em ICLR}, 2022.

\bibitem{luu2022utilizing}
T.~M. Luu, T.~Nguyen, T.~Vu, and C.~D. Yoo, ``Utilizing skipped frames in
  action repeats for improving sample efficiency in reinforcement learning,''
  {\em IEEE Access}, vol.~10, pp.~64965--64975, 2022.

\bibitem{huang2017adversarial}
S.~Huang, N.~Papernot, I.~Goodfellow, Y.~Duan, and P.~Abbeel, ``Adversarial
  attacks on neural network policies,'' {\em arXiv preprint arXiv:1702.02284},
  2017.

\bibitem{lin2017tactics}
Y.-C. Lin, Z.-W. Hong, Y.-H. Liao, M.-L. Shih, M.-Y. Liu, and M.~Sun, ``Tactics
  of adversarial attack on deep reinforcement learning agents,'' {\em arXiv
  preprint arXiv:1703.06748}, 2017.

\bibitem{kos2017delving}
J.~Kos and D.~Song, ``Delving into adversarial attacks on deep policies,'' {\em
  arXiv preprint arXiv:1705.06452}, 2017.

\bibitem{behzadan2017vulnerability}
V.~Behzadan and A.~Munir, ``Vulnerability of deep reinforcement learning to
  policy induction attacks,'' in {\em MLDM}, pp.~262--275, 2017.

\bibitem{pattanaik2018robust}
A.~Pattanaik, Z.~Tang, S.~Liu, G.~Bommannan, and G.~Chowdhary, ``Robust deep
  reinforcement learning with adversarial attacks,'' {\em AAMAS}, 2018.

\bibitem{shalev2016safe}
S.~Shalev-Shwartz, S.~Shammah, and A.~Shashua, ``Safe, multi-agent,
  reinforcement learning for autonomous driving,'' {\em arXiv preprint
  arXiv:1610.03295}, 2016.

\bibitem{sallab2017deep}
A.~E. Sallab, M.~Abdou, E.~Perot, and S.~Yogamani, ``Deep reinforcement
  learning framework for autonomous driving,'' {\em arXiv preprint
  arXiv:1704.02532}, 2017.

\bibitem{pan2017virtual}
X.~Pan, Y.~You, Z.~Wang, and C.~Lu, ``Virtual to real reinforcement learning
  for autonomous driving,'' {\em arXiv preprint arXiv:1704.03952}, 2017.

\bibitem{Gray_2019}
D.~Gray, ``Introducing voyage deepdrive - unlocking the potential of deep
  reinforcement learning,'' Nov 2019.

\bibitem{you2019advanced}
C.~You, J.~Lu, D.~Filev, and P.~Tsiotras, ``Advanced planning for autonomous
  vehicles using reinforcement learning and deep inverse reinforcement
  learning,'' {\em Robotics and Autonomous Systems}, vol.~114, pp.~1--18, 2019.

\bibitem{liang2022efficient}
Y.~Liang, Y.~Sun, R.~Zheng, and F.~Huang, ``Efficient adversarial training
  without attacking: Worst-case-aware robust reinforcement learning,'' {\em
  NeurIPS}, 2022.

\bibitem{he2022robust}
X.~He, B.~Lou, H.~Yang, and C.~Lv, ``Robust decision making for autonomous
  vehicles at highway on-ramps: A constrained adversarial reinforcement
  learning approach,'' {\em IEEE Transactions on Intelligent Transportation
  Systems}, vol.~24, no.~4, pp.~4103--4113, 2022.

\bibitem{zhang2020robust}
H.~Zhang, H.~Chen, C.~Xiao, B.~Li, M.~Liu, D.~Boning, and C.-J. Hsieh, ``Robust
  deep reinforcement learning against adversarial perturbations on state
  observations,'' {\em NeurIPS}, pp.~21024--21037, 2020.

\bibitem{zhang2021robust}
H.~Zhang, H.~Chen, D.~Boning, and C.-J. Hsieh, ``Robust reinforcement learning
  on state observations with learned optimal adversary,'' {\em ICLR}, 2021.

\bibitem{gupta2022rsac}
S.~Gupta, G.~Singal, D.~Garg, and S.~Das, ``Rsac: a robust deep reinforcement
  learning strategy for dimensionality perturbation,'' {\em IEEE Transactions
  on Emerging Topics in Computational Intelligence}, vol.~6, no.~5,
  pp.~1157--1166, 2022.

\bibitem{wu2022robust}
J.~Wu and Y.~Vorobeychik, ``Robust deep reinforcement learning through
  bootstrapped opportunistic curriculum,'' in {\em International Conference on
  Machine Learning}, pp.~24177--24211, PMLR, 2022.

\bibitem{liu2022robustness}
Z.~Liu, Z.~Guo, Z.~Cen, H.~Zhang, J.~Tan, B.~Li, and D.~Zhao, ``On the
  robustness of safe reinforcement learning under observational
  perturbations,'' {\em ICLR}, 2023.

\bibitem{korkmaz2023detecting}
E.~Korkmaz and J.~Brown-Cohen, ``Detecting adversarial directions in deep
  reinforcement learning to make robust decisions,'' {\em ICML}, 2023.

\bibitem{shen2020deep}
Q.~Shen, Y.~Li, H.~Jiang, Z.~Wang, and T.~Zhao, ``Deep reinforcement learning
  with robust and smooth policy,'' in {\em ICML}, pp.~8707--8718, 2020.

\bibitem{oikarinen2021robust}
T.~Oikarinen, W.~Zhang, A.~Megretski, L.~Daniel, and T.-W. Weng, ``Robust deep
  reinforcement learning through adversarial loss,'' {\em NeurIPS},
  pp.~26156--26167, 2021.

\bibitem{yang2022rorl}
R.~Yang, C.~Bai, X.~Ma, Z.~Wang, C.~Zhang, and L.~Han, ``Rorl: Robust offline
  reinforcement learning via conservative smoothing,'' in {\em NeurIPS}, 2022.

\bibitem{behzadan2017whatever}
V.~Behzadan and A.~Munir, ``Whatever does not kill deep reinforcement learning,
  makes it stronger,'' {\em arXiv preprint arXiv:1712.09344}, 2017.

\bibitem{sun2021strongest}
Y.~Sun, R.~Zheng, Y.~Liang, and F.~Huang, ``Who is the strongest enemy? towards
  optimal and efficient evasion attacks in deep rl,'' {\em ICLR}, 2022.

\bibitem{dziugaite2016study}
G.~K. Dziugaite, Z.~Ghahramani, and D.~M. Roy, ``A study of the effect of jpg
  compression on adversarial images,'' {\em arXiv:1608.00853}, 2016.

\bibitem{meng2017magnet}
D.~Meng and H.~Chen, ``Magnet: a two-pronged defense against adversarial
  examples,'' in {\em CCS}, pp.~135--147, 2017.

\bibitem{lu2017no}
J.~Lu, H.~Sibai, E.~Fabry, and D.~Forsyth, ``No need to worry about adversarial
  examples in object detection in autonomous vehicles,'' {\em
  arXiv:1707.03501}, 2017.

\bibitem{liao2018defense}
F.~Liao, M.~Liang, Y.~Dong, T.~Pang, X.~Hu, and J.~Zhu, ``Defense against
  adversarial attacks using high-level representation guided denoiser,'' in
  {\em CVPR}, pp.~1778--1787, 2018.

\bibitem{guo2018countering}
C.~Guo, M.~Rana, M.~Cisse, and L.~Van Der~Maaten, ``Countering adversarial
  images using input transformations,'' {\em ICLR}, 2018.

\bibitem{xu2018feature}
W.~Xu, D.~Evans, and Y.~Qi, ``Feature squeezing: Detecting adversarial examples
  in deep neural networks,'' {\em NDSS}, 2018.

\bibitem{prakash2018deflecting}
A.~Prakash, N.~Moran, S.~Garber, A.~DiLillo, and J.~Storer, ``Deflecting
  adversarial attacks with pixel deflection,'' in {\em CVPR}, pp.~8571--8580,
  2018.

\bibitem{samangouei2018defense}
P.~Samangouei, M.~Kabkab, and R.~Chellappa, ``Defense-gan: Protecting
  classifiers against adversarial attacks using generative models,'' {\em
  ICLR}, 2018.

\bibitem{gupta2019ciidefence}
P.~Gupta and E.~Rahtu, ``Ciidefence: Defeating adversarial attacks by fusing
  class-specific image inpainting and image denoising,'' in {\em Proceedings of
  the IEEE/CVF International Conference on Computer Vision}, pp.~6708--6717,
  2019.

\bibitem{jin2019ape}
G.~Jin, S.~Shen, D.~Zhang, F.~Dai, and Y.~Zhang, ``Ape-gan: Adversarial
  perturbation elimination with gan,'' in {\em ICASSP}, pp.~3842--3846, IEEE,
  2019.

\bibitem{salman2020denoised}
H.~Salman, M.~Sun, G.~Yang, A.~Kapoor, and J.~Z. Kolter, ``Denoised smoothing:
  A provable defense for pretrained classifiers,'' {\em Advances in Neural
  Information Processing Systems}, vol.~33, pp.~21945--21957, 2020.

\bibitem{nie2022diffusion}
W.~Nie, B.~Guo, Y.~Huang, C.~Xiao, A.~Vahdat, and A.~Anandkumar, ``Diffusion
  models for adversarial purification,'' {\em ICML}, 2022.

\bibitem{szegedy2013intriguing}
C.~Szegedy, W.~Zaremba, I.~Sutskever, J.~Bruna, D.~Erhan, I.~Goodfellow, and
  R.~Fergus, ``Intriguing properties of neural networks,'' {\em arXiv preprint
  arXiv:1312.6199}, 2013.

\bibitem{goodfellow2014explaining}
I.~J. Goodfellow, J.~Shlens, and C.~Szegedy, ``Explaining and harnessing
  adversarial examples,'' {\em arXiv:1412.6572}, 2014.

\bibitem{sun2020stealthy}
J.~Sun, T.~Zhang, X.~Xie, L.~Ma, Y.~Zheng, K.~Chen, and Y.~Liu, ``Stealthy and
  efficient adversarial attacks against deep reinforcement learning,'' in {\em
  Proceedings of the AAAI Conference on Artificial Intelligence}, vol.~34,
  pp.~5883--5891, 2020.

\bibitem{mo2022attacking}
K.~Mo, W.~Tang, J.~Li, and X.~Yuan, ``Attacking deep reinforcement learning
  with decoupled adversarial policy,'' {\em IEEE Transactions on Dependable and
  Secure Computing}, vol.~20, no.~1, pp.~758--768, 2022.

\bibitem{russo2021optimal}
A.~Russo and A.~Proutiere, ``Optimal attacks on reinforcement learning
  policies,'' {\em ACC}, 2021.

\bibitem{goodfellow2014generative}
I.~J. Goodfellow, J.~Pouget-Abadie, M.~Mirza, B.~Xu, D.~Warde-Farley, S.~Ozair,
  A.~Courville, and Y.~Bengio, ``Generative adversarial networks,'' 2014.

\bibitem{ho2020denoising}
J.~Ho, A.~Jain, and P.~Abbeel, ``Denoising diffusion probabilistic models,''
  {\em arXiv preprint arxiv:2006.11239}, 2020.

\bibitem{schulman2017proximal}
J.~Schulman, F.~Wolski, P.~Dhariwal, A.~Radford, and O.~Klimov, ``Proximal
  policy optimization algorithms,'' {\em arXiv preprint arXiv:1707.06347},
  2017.

\bibitem{weng2019toward}
T.-W. Weng, K.~D. Dvijotham, J.~Uesato, K.~Xiao, S.~Gowal, R.~Stanforth, and
  P.~Kohli, ``Toward evaluating robustness of deep reinforcement learning with
  continuous control,'' in {\em International Conference on Learning
  Representations}, 2019.

\bibitem{fischer2019online}
M.~Fischer, M.~Mirman, S.~Stalder, and M.~Vechev, ``Online robustness training
  for deep reinforcement learning,'' {\em arXiv preprint arXiv:1911.00887},
  2019.

\bibitem{achiam2017constrained}
J.~Achiam, D.~Held, A.~Tamar, and P.~Abbeel, ``Constrained policy
  optimization,'' in {\em International conference on machine learning},
  pp.~22--31, PMLR, 2017.

\bibitem{bengio2013estimating}
Y.~Bengio, N.~L{\'e}onard, and A.~Courville, ``Estimating or propagating
  gradients through stochastic neurons for conditional computation,'' {\em
  arXiv:1308.3432}, 2013.

\bibitem{brockman2016openai}
G.~Brockman, V.~Cheung, L.~Pettersson, J.~Schneider, J.~Schulman, J.~Tang, and
  W.~Zaremba, ``Openai gym,'' {\em arXiv preprint arXiv:1606.01540}, 2016.

\bibitem{lillicrap2015continuous}
T.~P. Lillicrap, J.~J. Hunt, A.~Pritzel, N.~Heess, T.~Erez, Y.~Tassa,
  D.~Silver, and D.~Wierstra, ``Continuous control with deep reinforcement
  learning,'' {\em ICML}, 2016.

\bibitem{d3rlpy}
T.~Seno and M.~Imai, ``d3rlpy: An offline deep reinforcement learning
  library,'' {\em Journal of Machine Learning Research}, vol.~23, no.~315,
  pp.~1--20, 2022.

\bibitem{kurakin2016adversarial}
A.~Kurakin, I.~Goodfellow, and S.~Bengio, ``Adversarial machine learning at
  scale,'' {\em arXiv preprint arXiv:1611.01236}, 2016.

\bibitem{kingma2014method}
D.~P. Kingma and J.~Ba, ``Adam: A method for stochastic optimization,'' {\em
  Proc. Int. Conf. Learn. Represent. (ICLR)}, pp.~1--15, 2014.

\bibitem{kingma2013auto}
D.~P. Kingma and M.~Welling, ``Auto-encoding variational bayes,'' {\em arXiv
  preprint arXiv:1312.6114}, 2013.

\end{thebibliography}




@article{gupta2022rsac,
  title={RSAC: a robust deep reinforcement learning strategy for dimensionality perturbation},
  author={Gupta, Surbhi and Singal, Gaurav and Garg, Deepak and Das, Swagatam},
  journal={IEEE Transactions on Emerging Topics in Computational Intelligence},
  volume={6},
  number={5},
  pages={1157--1166},
  year={2022},
  publisher={IEEE}
}




@article{mnih2015human,
	title={Human-level control through deep reinforcement learning},
	author={Mnih, Volodymyr and Kavukcuoglu, Koray and Silver, David and Rusu, Andrei A and Veness, Joel and Bellemare, Marc G and Graves, Alex and Riedmiller, Martin and Fidjeland, Andreas K and Ostrovski, Georg and others},
	journal={Nature},
	pages={529-533},
	year={2015}
}

@inproceedings{schulman2015trust,
	title={Trust region policy optimization},
	author={Schulman, John and Levine, Sergey and Abbeel, Pieter and Jordan, Michael and Moritz, Philipp},
	booktitle={ICML},
	pages={1889--1897},
	year={2015}
}

@article{lillicrap2015continuous,
	title={Continuous control with deep reinforcement learning},
	author={Lillicrap, Timothy P and Hunt, Jonathan J and Pritzel, Alexander and Heess, Nicolas and Erez, Tom and Tassa, Yuval and Silver, David and Wierstra, Daan},
	journal={ICML},
	year={2016}
}

@article{silver2016mastering,
	title={Mastering the game of Go with deep neural networks and tree search},
	author={Silver, David and Huang, Aja and Maddison, Chris J and Guez, Arthur and Sifre, Laurent and Van Den Driessche, George and Schrittwieser, Julian and Antonoglou, Ioannis and Panneershelvam, Veda and Lanctot, Marc and others},
	journal={Nature},
	pages={484--489},
	year={2016},
}

@inproceedings{fujimoto2018addressing,
	title={Addressing function approximation error in actor-critic methods},
	author={Fujimoto, Scott and Hoof, Herke and Meger, David},
	booktitle={ICML},
	pages={1587--1596},
	year={2018}
}

@article{schulman2017proximal,
	title={Proximal policy optimization algorithms},
	author={Schulman, John and Wolski, Filip and Dhariwal, Prafulla and Radford, Alec and Klimov, Oleg},
	journal={arXiv preprint arXiv:1707.06347},
	year={2017}
}

@inproceedings{haarnoja2018soft,
	title={Soft actor-critic: Off-policy maximum entropy deep reinforcement learning with a stochastic actor},
	author={Haarnoja, Tuomas and Zhou, Aurick and Abbeel, Pieter and Levine, Sergey},
	booktitle={ICML},
	pages={1861--1870},
	year={2018}
}

@article{pattanaik2018robust,
	title={Robust deep reinforcement learning with adversarial attacks},
	author={Pattanaik, Anay and Tang, Zhenyi and Liu, Shuijing and Bommannan, Gautham and Chowdhary, Girish},
	journal={AAMAS},
	year={2018}
}

@article{huang2017adversarial,
	title={Adversarial attacks on neural network policies},
	author={Huang, Sandy and Papernot, Nicolas and Goodfellow, Ian and Duan, Yan and Abbeel, Pieter},
	journal={arXiv preprint arXiv:1702.02284},
	year={2017}
}

@article{lin2017tactics,
	title={Tactics of adversarial attack on deep reinforcement learning agents},
	author={Lin, Yen-Chen and Hong, Zhang-Wei and Liao, Yuan-Hong and Shih, Meng-Li and Liu, Ming-Yu and Sun, Min},
	journal={arXiv preprint arXiv:1703.06748},
	year={2017}
}

@inproceedings{behzadan2017vulnerability,
	title={Vulnerability of deep reinforcement learning to policy induction attacks},
	author={Behzadan, Vahid and Munir, Arslan},
	booktitle={MLDM},
	pages={262--275},
	year={2017}
}

@article{behzadan2017whatever,
	title={Whatever does not kill deep reinforcement learning, makes it stronger},
	author={Behzadan, Vahid and Munir, Arslan},
	journal={arXiv preprint arXiv:1712.09344},
	year={2017}
}

@article{kos2017delving,
	title={Delving into adversarial attacks on deep policies},
	author={Kos, Jernej and Song, Dawn},
	journal={arXiv preprint arXiv:1705.06452},
	year={2017}
}

@article{sallab2017deep,
	title={Deep reinforcement learning framework for autonomous driving},
	author={Sallab, Ahmad EL and Abdou, Mohammed and Perot, Etienne and Yogamani, Senthil},
	journal={arXiv preprint arXiv:1704.02532},
	year={2017}
}

@article{pan2017virtual,
	title={Virtual to real reinforcement learning for autonomous driving},
	author={Pan, Xinlei and You, Yurong and Wang, Ziyan and Lu, Cewu},
	journal={arXiv preprint arXiv:1704.03952},
	year={2017}
}

@inproceedings{shen2020deep,
	title={Deep reinforcement learning with robust and smooth policy},
	author={Shen, Qianli and Li, Yan and Jiang, Haoming and Wang, Zhaoran and Zhao, Tuo},
	booktitle={ICML},
	pages={8707--8718},
	year={2020},
}

@article{russo2021optimal,
  title={Optimal attacks on reinforcement learning policies},
  author={Russo, Alessio and Proutiere, Alexandre},
  journal={ACC},
  year={2021}
}

@misc{Gray_2019, title={Introducing voyage deepdrive - unlocking the potential of deep reinforcement learning}, url={https://news.voyage.auto/introducing-voyage-deepdrive-69b3cf0f0be6}, journal={Medium}, publisher={Voyage}, author={Gray, Drew}, year={2019}, month={Nov}} 

@article{shalev2016safe,
	title={Safe, multi-agent, reinforcement learning for autonomous driving},
	author={Shalev-Shwartz, Shai and Shammah, Shaked and Shashua, Amnon},
	journal={arXiv preprint arXiv:1610.03295},
	year={2016}
}

@article{you2019advanced,
	title={Advanced planning for autonomous vehicles using reinforcement learning and deep inverse reinforcement learning},
	author={You, Changxi and Lu, Jianbo and Filev, Dimitar and Tsiotras, Panagiotis},
	journal={Robotics and Autonomous Systems},
	volume={114},
	pages={1--18},
	year={2019},
	publisher={Elsevier}
}

@article{zhang2020robust,
	title={Robust deep reinforcement learning against adversarial perturbations on state observations},
	author={Zhang, Huan and Chen, Hongge and Xiao, Chaowei and Li, Bo and Liu, Mingyan and Boning, Duane and Hsieh, Cho-Jui},
	journal={NeurIPS},
	pages={21024--21037},
	year={2020}
}

@article{oikarinen2021robust,
	title={Robust deep reinforcement learning through adversarial loss},
	author={Oikarinen, Tuomas and Zhang, Wang and Megretski, Alexandre and Daniel, Luca and Weng, Tsui-Wei},
	journal={NeurIPS},
	pages={26156--26167},
	year={2021}
}

@inproceedings{yang2022rorl,
	title={RORL: Robust Offline Reinforcement Learning via Conservative Smoothing},
	author={Yang, Rui and Bai, Chenjia and Ma, Xiaoteng and Wang, Zhaoran and Zhang, Chongjie and Han, Lei},
	booktitle={NeurIPS},
	year={2022}
}

@article{zhang2021robust,
	title={Robust reinforcement learning on state observations with learned optimal adversary},
	author={Zhang, Huan and Chen, Hongge and Boning, Duane and Hsieh, Cho-Jui},
	journal={ICLR},
	year={2021}
}

@article{sun2021strongest,
	title={Who is the strongest enemy? towards optimal and efficient evasion attacks in deep rl},
	author={Sun, Yanchao and Zheng, Ruijie and Liang, Yongyuan and Huang, Furong},
	journal={ICLR},
	year={2022}
}

@article{liang2022efficient,
	title={Efficient Adversarial Training without Attacking: Worst-Case-Aware Robust Reinforcement Learning},
	author={Liang, Yongyuan and Sun, Yanchao and Zheng, Ruijie and Huang, Furong},
	journal={NeurIPS},
	year={2022}
}

@inproceedings{fujimoto2019off,
	title={Off-Policy Deep Reinforcement Learning without Exploration},
	author={Fujimoto, Scott and Meger, David and Precup, Doina},
	booktitle={ICML},
	pages={2052--2062},
	year={2019}
}

@article{kumar2019stabilizing,
	title={Stabilizing off-policy q-learning via bootstrapping error reduction},
	author={Kumar, Aviral and Fu, Justin and Soh, Matthew and Tucker, George and Levine, Sergey},
	journal={NeurIPS},
	volume={32},
	year={2019}
}

@article{fujimoto2021minimalist,
	title={A minimalist approach to offline reinforcement learning},
	author={Fujimoto, Scott and Gu, Shixiang Shane},
	journal={NeurIPS},
	volume={34},
	pages={20132--20145},
	year={2021}
}

@article{kumar2020conservative,
	title={Conservative q-learning for offline reinforcement learning},
	author={Kumar, Aviral and Zhou, Aurick and Tucker, George and Levine, Sergey},
	journal={NeurIPS},
	volume={33},
	pages={1179--1191},
	year={2020}
}

@article{bai2022pessimistic,
	title={Pessimistic bootstrapping for uncertainty-driven offline reinforcement learning},
	author={Bai, Chenjia and Wang, Lingxiao and Yang, Zhuoran and Deng, Zhihong and Garg, Animesh and Liu, Peng and Wang, Zhaoran},
	journal={ICLR},
	year={2022}
}

@inproceedings{meng2017magnet,
	title={Magnet: a two-pronged defense against adversarial examples},
	author={Meng, Dongyu and Chen, Hao},
	booktitle={CCS},
	pages={135--147},
	year={2017}
}

@article{guo2018countering,
	title={Countering adversarial images using input transformations},
	author={Guo, Chuan and Rana, Mayank and Cisse, Moustapha and Van Der Maaten, Laurens},
	journal={ICLR},
	year={2018}
}

@inproceedings{liao2018defense,
	title={Defense against adversarial attacks using high-level representation guided denoiser},
	author={Liao, Fangzhou and Liang, Ming and Dong, Yinpeng and Pang, Tianyu and Hu, Xiaolin and Zhu, Jun},
	booktitle={CVPR},
	pages={1778--1787},
	year={2018}
}

@inproceedings{gupta2019ciidefence,
	title={Ciidefence: Defeating adversarial attacks by fusing class-specific image inpainting and image denoising},
	author={Gupta, Puneet and Rahtu, Esa},
	booktitle={Proceedings of the IEEE/CVF International Conference on Computer Vision},
	pages={6708--6717},
	year={2019}
}

@article{samangouei2018defense,
	title={Defense-gan: Protecting classifiers against adversarial attacks using generative models},
	author={Samangouei, Pouya and Kabkab, Maya and Chellappa, Rama},
	journal={ICLR},
	year={2018}
}

@inproceedings{jin2019ape,
	title={Ape-gan: Adversarial perturbation elimination with gan},
	author={Jin, Guoqing and Shen, Shiwei and Zhang, Dongming and Dai, Feng and Zhang, Yongdong},
	booktitle={ICASSP},
	pages={3842--3846},
	year={2019},
	organization={IEEE}
}

@article{salman2020denoised,
	title={Denoised smoothing: A provable defense for pretrained classifiers},
	author={Salman, Hadi and Sun, Mingjie and Yang, Greg and Kapoor, Ashish and Kolter, J Zico},
	journal={Advances in Neural Information Processing Systems},
	volume={33},
	pages={21945--21957},
	year={2020}
}

@article{nie2022diffusion,
	title={Diffusion models for adversarial purification},
	author={Nie, Weili and Guo, Brandon and Huang, Yujia and Xiao, Chaowei and Vahdat, Arash and Anandkumar, Anima},
	journal={ICML},
	year={2022}
}


@article{xu2018feature,
	title={Feature squeezing: Detecting adversarial examples in deep neural networks},
	author={Xu, Weilin and Evans, David and Qi, Yanjun},
	journal={NDSS},
	year={2018}
}

@inproceedings{prakash2018deflecting,
	title={Deflecting adversarial attacks with pixel deflection},
	author={Prakash, Aaditya and Moran, Nick and Garber, Solomon and DiLillo, Antonella and Storer, James},
	booktitle={CVPR},
	pages={8571--8580},
	year={2018}
}

@article{dziugaite2016study,
	title={A study of the effect of jpg compression on adversarial images},
	author={Dziugaite, Gintare Karolina and Ghahramani, Zoubin and Roy, Daniel M},
	journal={arXiv:1608.00853},
	year={2016}
}

@article{lu2017no,
	title={No need to worry about adversarial examples in object detection in autonomous vehicles},
	author={Lu, Jiajun and Sibai, Hussein and Fabry, Evan and Forsyth, David},
	journal={arXiv:1707.03501},
	year={2017}
}

@article{van2017neural,
	title={Neural discrete representation learning},
	author={Van Den Oord, Aaron and Vinyals, Oriol and others},
	journal={Advances in neural information processing systems},
	volume={30},
	year={2017}
}


@article{razavi2019generating,
	title={Generating diverse high-fidelity images with vq-vae-2},
	author={Razavi, Ali and Van den Oord, Aaron and Vinyals, Oriol},
	journal={Advances in neural information processing systems},
	volume={32},
	year={2019}
}

@article{szegedy2013intriguing,
	title={Intriguing properties of neural networks},
	author={Szegedy, Christian and Zaremba, Wojciech and Sutskever, Ilya and Bruna, Joan and Erhan, Dumitru and Goodfellow, Ian and Fergus, Rob},
	journal={arXiv preprint arXiv:1312.6199},
	year={2013}
}

@article{baevski2019vq,
	title={vq-wav2vec: Self-supervised learning of discrete speech representations},
	author={Baevski, Alexei and Schneider, Steffen and Auli, Michael},
	journal={arXiv preprint arXiv:1910.05453},
	year={2019}
}

@inproceedings{esser2021taming,
	title={Taming transformers for high-resolution image synthesis},
	author={Esser, Patrick and Rombach, Robin and Ommer, Bjorn},
	booktitle={Proceedings of the IEEE/CVF conference on computer vision and pattern recognition},
	pages={12873--12883},
	year={2021}
}

@inproceedings{ozair2021vector,
	title={Vector quantized models for planning},
	author={Ozair, Sherjil and Li, Yazhe and Razavi, Ali and Antonoglou, Ioannis and Van Den Oord, Aaron and Vinyals, Oriol},
	booktitle={ICML},
	pages={8302--8313},
	year={2021},
	organization={PMLR}
}


@article{bengio2013estimating,
	title={Estimating or propagating gradients through stochastic neurons for conditional computation},
	author={Bengio, Yoshua and L{\'e}onard, Nicholas and Courville, Aaron},
	journal={arXiv:1308.3432},
	year={2013}
}

@article{goodfellow2014explaining,
	title={Explaining and harnessing adversarial examples},
	author={Goodfellow, Ian J and Shlens, Jonathon and Szegedy, Christian},
	journal={arXiv:1412.6572},
	year={2014}
}

@misc{goodfellow2014generative,
      title={Generative Adversarial Networks}, 
      author={Ian J. Goodfellow and Jean Pouget-Abadie and Mehdi Mirza and Bing Xu and David Warde-Farley and Sherjil Ozair and Aaron Courville and Yoshua Bengio},
      year={2014},
      eprint={1406.2661},
      archivePrefix={arXiv},
      primaryClass={stat.ML}
}

@article{ho2020denoising,
  title={Denoising Diffusion Probabilistic Models},
  author={Jonathan Ho and Ajay Jain and Pieter Abbeel},
  year={2020},
  journal={arXiv preprint arxiv:2006.11239}
}

@article{iyengar2005robust,
	title={Robust dynamic programming},
	author={Iyengar, Garud N},
	journal={Mathematics of Operations Research},
	volume={30},
	number={2},
	pages={257--280},
	year={2005},
	publisher={INFORMS}
}

@article{nilim2003robustness,
	title={Robustness in markov decision problems with uncertain transition matrices},
	author={Nilim, Arnab and Ghaoui, Laurent},
	journal={Advances in neural information processing systems},
	volume={16},
	year={2003}
}

@article{mankowitz2020robust,
	title={Robust reinforcement learning for continuous control with model misspecification},
	author={Mankowitz, Daniel J and Levine, Nir and Jeong, Rae and Shi, Yuanyuan and Kay, Jackie and Abdolmaleki, Abbas and Springenberg, Jost Tobias and Mann, Timothy and Hester, Todd and Riedmiller, Martin},
	journal={ICLR},
	year={2020}
}

@article{qu2020defending,
	title={Defending adversarial attacks without adversarial attacks in deep reinforcement learning},
	author={Qu, Xinghua and Ong, Yew-Soon and Gupta, Abhishek and Sun, Zhu},
	journal={arXiv:2008.06199},
	year={2020}
}


@article{sadeghi2016cad2rl,
	title={Cad2rl: Real single-image flight without a single real image},
	author={Sadeghi, Fereshteh and Levine, Sergey},
	journal={arXiv preprint arXiv:1611.04201},
	year={2016}
}

@inproceedings{tobin2017domain,
	title={Domain randomization for transferring deep neural networks from simulation to the real world},
	author={Tobin, Josh and Fong, Rachel and Ray, Alex and Schneider, Jonas and Zaremba, Wojciech and Abbeel, Pieter},
	booktitle={2017 IEEE/RSJ international conference on intelligent robots and systems (IROS)},
	pages={23--30},
	year={2017},
	organization={IEEE}
}

@article{pinto2017asymmetric,
	title={Asymmetric actor critic for image-based robot learning},
	author={Pinto, Lerrel and Andrychowicz, Marcin and Welinder, Peter and Zaremba, Wojciech and Abbeel, Pieter},
	journal={arXiv preprint arXiv:1710.06542},
	year={2017}
}

@article{lee2020network,
	title={Network randomization: A simple technique for generalization in deep reinforcement learning},
	author={Lee, Kimin and Lee, Kibok and Shin, Jinwoo and Lee, Honglak},
	journal={ICLR},
	year={2020}
}

@article{laskin2020reinforcement,
	title={Reinforcement learning with augmented data},
	author={Laskin, Misha and Lee, Kimin and Stooke, Adam and Pinto, Lerrel and Abbeel, Pieter and Srinivas, Aravind},
	journal={Advances in neural information processing systems},
	volume={33},
	pages={19884--19895},
	year={2020}
}

@inproceedings{cobbe2019quantifying,
	title={Quantifying generalization in reinforcement learning},
	author={Cobbe, Karl and Klimov, Oleg and Hesse, Chris and Kim, Taehoon and Schulman, John},
	booktitle={International Conference on Machine Learning},
	pages={1282--1289},
	year={2019},
	organization={PMLR}
}

@article{kostrikov2020image,
	title={Image augmentation is all you need: Regularizing deep reinforcement learning from pixels},
	author={Kostrikov, Ilya and Yarats, Denis and Fergus, Rob},
	journal={arXiv preprint arXiv:2004.13649},
	year={2020}
}

@article{fu2020d4rl,
	title={D4rl: Datasets for deep data-driven reinforcement learning},
	author={Fu, Justin and Kumar, Aviral and Nachum, Ofir and Tucker, George and Levine, Sergey},
	journal={arXiv:2004.07219},
	year={2020}
}

@article{kurakin2016adversarial,
	title={Adversarial machine learning at scale},
	author={Kurakin, Alexey and Goodfellow, Ian and Bengio, Samy},
	journal={arXiv preprint arXiv:1611.01236},
	year={2016}
}

@article{liu2021discrete,
	title={Discrete-valued neural communication},
	author={Liu, Dianbo and Lamb, Alex M and Kawaguchi, Kenji and ALIAS PARTH GOYAL, Anirudh Goyal and Sun, Chen and Mozer, Michael C and Bengio, Yoshua},
	journal={Advances in Neural Information Processing Systems},
	volume={34},
	pages={2109--2121},
	year={2021}
}

@article{islam2022discrete,
	title={Discrete Factorial Representations as an Abstraction for Goal Conditioned Reinforcement Learning},
	author={Islam, Riashat and Zang, Hongyu and Goyal, Anirudh and Lamb, Alex and Kawaguchi, Kenji and Li, Xin and Laroche, Romain and Bengio, Yoshua and Combes, Remi Tachet Des},
	journal={arXiv preprint arXiv:2211.00247},
	year={2022}
}

@inproceedings{giannakopoulos2021neural,
	title={Neural discrete abstraction of high-dimensional spaces: A case study in reinforcement learning},
	author={Giannakopoulos, Petros and Pikrakis, Aggelos and Cotronis, Yannis},
	booktitle={2020 28th European Signal Processing Conference (EUSIPCO)},
	pages={1517--1521},
	year={2021},
	organization={IEEE}
}

@article{micheli2023transformers,
	title={Transformers are sample efficient world models},
	author={Micheli, Vincent and Alonso, Eloi and Fleuret, Fran{\c{c}}ois},
	journal={ICLR},
	year={2023}
}

@article{bottou1994convergence,
	title={Convergence properties of the k-means algorithms},
	author={Bottou, Leon and Bengio, Yoshua},
	journal={Advances in neural information processing systems},
	volume={7},
	year={1994}
}

@article{d3rlpy,
	author  = {Takuma Seno and Michita Imai},
	title   = {d3rlpy: An Offline Deep Reinforcement Learning Library},
	journal = {Journal of Machine Learning Research},
	year    = {2022},
	volume  = {23},
	number  = {315},
	pages   = {1--20},
	url     = {http://jmlr.org/papers/v23/22-0017.html}
}

@article{madry2017towards,
	title={Towards deep learning models resistant to adversarial attacks},
	author={Madry, Aleksander and Makelov, Aleksandar and Schmidt, Ludwig and Tsipras, Dimitris and Vladu, Adrian},
	journal={arXiv preprint arXiv:1706.06083},
	year={2017}
}

@inproceedings{zhang2019theoretically,
	title={Theoretically principled trade-off between robustness and accuracy},
	author={Zhang, Hongyang and Yu, Yaodong and Jiao, Jiantao and Xing, Eric and El Ghaoui, Laurent and Jordan, Michael},
	booktitle={International conference on machine learning},
	pages={7472--7482},
	year={2019},
	organization={PMLR}
}

@inproceedings{wang2020improving,
	title={Improving adversarial robustness requires revisiting misclassified examples},
	author={Wang, Yisen and Zou, Difan and Yi, Jinfeng and Bailey, James and Ma, Xingjun and Gu, Quanquan},
	booktitle={International Conference on Learning Representations},
	year={2020}
}

@article{ding2020mma,
	title={Mma training: Direct input space margin maximization through adversarial training},
	author={Ding, Gavin Weiguang and Sharma, Yash and Lui, Kry Yik Chau and Huang, Ruitong},
	journal={ICLR},
	year={2020}
}

@article{levine2020offline,
	title={Offline reinforcement learning: Tutorial, review, and perspectives on open problems},
	author={Levine, Sergey and Kumar, Aviral and Tucker, George and Fu, Justin},
	journal={arXiv preprint arXiv:2005.01643},
	year={2020}
}

@inproceedings{fernandez2000vqql,
	title={VQQL. Applying vector quantization to reinforcement learning},
	author={Fern{\'a}ndez, Fernando and Borrajo, Daniel},
	booktitle={RoboCup-99: Robot Soccer World Cup III 3},
	pages={292--303},
	year={2000},
	organization={Springer}
}

@inproceedings{mavridis2021vector,
	title={Vector quantization for adaptive state aggregation in reinforcement learning},
	author={Mavridis, Christos N and Baras, John S},
	booktitle={2021 American Control Conference (ACC)},
	pages={2187--2192},
	year={2021},
	organization={IEEE}
}

@article{lau2002adaptive,
	title={Adaptive vector quantization for reinforcement learning},
	author={Lau, HYK and Mak, KL and Lee, ISK},
	journal={IFAC Proceedings Volumes},
	volume={35},
	number={1},
	pages={493--498},
	year={2002},
	publisher={Elsevier}
}

@inproceedings{van2016deep,
	title={Deep reinforcement learning with double q-learning},
	author={Van Hasselt, Hado and Guez, Arthur and Silver, David},
	booktitle={Proceedings of the AAAI conference on artificial intelligence},
	volume={30},
	number={1},
	year={2016}
}

@article{mnih2013playing,
	title={Playing atari with deep reinforcement learning},
	author={Mnih, Volodymyr and Kavukcuoglu, Koray and Silver, David and Graves, Alex and Antonoglou, Ioannis and Wierstra, Daan and Riedmiller, Martin},
	journal={arXiv preprint arXiv:1312.5602},
	year={2013}
}

@inproceedings{wang2016dueling,
	title={Dueling network architectures for deep reinforcement learning},
	author={Wang, Ziyu and Schaul, Tom and Hessel, Matteo and Hasselt, Hado and Lanctot, Marc and Freitas, Nando},
	booktitle={International conference on machine learning},
	pages={1995--2003},
	year={2016},
	organization={PMLR}
}

@article{kingma2013auto,
  title={Auto-encoding variational bayes},
  author={Kingma, Diederik P and Welling, Max},
  journal={arXiv preprint arXiv:1312.6114},
  year={2013}
}

@article{brockman2016openai,
  title={Openai gym},
  author={Brockman, Greg and Cheung, Vicki and Pettersson, Ludwig and Schneider, Jonas and Schulman, John and Tang, Jie and Zaremba, Wojciech},
  journal={arXiv preprint arXiv:1606.01540},
  year={2016}
}

@article{kingma2014method,
	title = {Adam: A Method for Stochastic Optimization},
	author = {Kingma, Diederik P. and Ba, Jimmy},
	journal={Proc. Int. Conf. Learn. Represent. (ICLR)},
	pages={1-15},
	year = 2014
}

@inproceedings{wu2022robust,
  title={Robust Deep Reinforcement Learning through Bootstrapped Opportunistic Curriculum},
  author={Wu, Junlin and Vorobeychik, Yevgeniy},
  booktitle={International Conference on Machine Learning},
  pages={24177--24211},
  year={2022},
  organization={PMLR}
}

@article{liu2022robustness,
  title={On the robustness of safe reinforcement learning under observational perturbations},
  author={Liu, Zuxin and Guo, Zijian and Cen, Zhepeng and Zhang, Huan and Tan, Jie and Li, Bo and Zhao, Ding},
  journal={ICLR},
  year={2023}
}

@article{korkmaz2023detecting,
  title={Detecting Adversarial Directions in Deep Reinforcement Learning to Make Robust Decisions},
  author={Korkmaz, Ezgi and Brown-Cohen, Jonah},
  journal={ICML},
  year={2023}
}

@inproceedings{weng2019toward,
  title={Toward evaluating robustness of deep reinforcement learning with continuous control},
  author={Weng, Tsui-Wei and Dvijotham, Krishnamurthy Dj and Uesato, Jonathan and Xiao, Kai and Gowal, Sven and Stanforth, Robert and Kohli, Pushmeet},
  booktitle={International Conference on Learning Representations},
  year={2019}
}

@article{wang2019verification,
  title={Verification of neural network control policy under persistent adversarial perturbation},
  author={Wang, Yuh-Shyang and Weng, Tsui-Wei and Daniel, Luca},
  journal={arXiv preprint arXiv:1908.06353},
  year={2019}
}

@article{fischer2019online,
  title={Online robustness training for deep reinforcement learning},
  author={Fischer, Marc and Mirman, Matthew and Stalder, Steven and Vechev, Martin},
  journal={arXiv preprint arXiv:1911.00887},
  year={2019}
}

@article{luu2021hindsight,
  title={Hindsight goal ranking on replay buffer for sparse reward environment},
  author={Luu, Tung M and Yoo, Chang D},
  journal={IEEE Access},
  volume={9},
  pages={51996--52007},
  year={2021},
  publisher={IEEE}
}

@inproceedings{achiam2017constrained,
  title={Constrained policy optimization},
  author={Achiam, Joshua and Held, David and Tamar, Aviv and Abbeel, Pieter},
  booktitle={International conference on machine learning},
  pages={22--31},
  year={2017},
  organization={PMLR}
}

@article{kostrikov2021offline,
  title={Offline reinforcement learning with implicit q-learning},
  author={Kostrikov, Ilya and Nair, Ashvin and Levine, Sergey},
  journal={ICLR},
  year={2022}
}

@article{mo2022attacking,
  title={Attacking deep reinforcement learning with decoupled adversarial policy},
  author={Mo, Kanghua and Tang, Weixuan and Li, Jin and Yuan, Xu},
  journal={IEEE Transactions on Dependable and Secure Computing},
  volume={20},
  number={1},
  pages={758--768},
  year={2022},
  publisher={IEEE}
}

@inproceedings{sun2020stealthy,
  title={Stealthy and efficient adversarial attacks against deep reinforcement learning},
  author={Sun, Jianwen and Zhang, Tianwei and Xie, Xiaofei and Ma, Lei and Zheng, Yan and Chen, Kangjie and Liu, Yang},
  booktitle={Proceedings of the AAAI Conference on Artificial Intelligence},
  volume={34},
  number={04},
  pages={5883--5891},
  year={2020}
}

@article{he2022robust,
  title={Robust decision making for autonomous vehicles at highway on-ramps: A constrained adversarial reinforcement learning approach},
  author={He, Xiangkun and Lou, Baichuan and Yang, Haohan and Lv, Chen},
  journal={IEEE Transactions on Intelligent Transportation Systems},
  volume={24},
  number={4},
  pages={4103--4113},
  year={2022},
  publisher={IEEE}
}

@article{luu2022utilizing,
  title={Utilizing skipped frames in action repeats for improving sample efficiency in reinforcement learning},
  author={Luu, Tung M and Nguyen, Thanh and Vu, Thang and Yoo, Chang D},
  journal={IEEE Access},
  volume={10},
  pages={64965--64975},
  year={2022},
  publisher={IEEE}
}

\EOD
\end{document}